\documentclass[runningheads,a4paper]{iscide}

\usepackage{amssymb}
\usepackage{graphicx}

\usepackage{times}

\usepackage{subfigure}
\usepackage{helvet}
\usepackage{courier}
\usepackage{amssymb}
\usepackage{wrapfig}
\usepackage{amsmath}
\usepackage{amsfonts,graphicx,algorithm,algorithmic,epstopdf}
\newcommand{\bs}[1]{\boldsymbol{#1}}

\graphicspath{{img/}}
\usepackage{url}
\urldef{\mailsa}\path|{alfred.hofmann, ursula.barth, ingrid.haas, frank.holzwarth,|
\urldef{\mailsb}\path|anna.kramer, leonie.kunz, christine.reiss, nicole.sator,|
\urldef{\mailsc}\path|erika.siebert-cole, peter.strasser}@iscide.edu|

\begin{document}

\mainmatter  

\title{Location Dependent Dirichlet Processes}

\titlerunning{}
%
\author{Shiliang Sun\inst{1}
\and John Paisley\inst{2} \and Qiuyang Liu\inst{1}
}
%
\institute{
Department of Computer Science and Technology, East China Normal University,\\
3663 North Zhongshan Road, Shanghai 200062, China\\
\and Department of Electrical Engineering, Columbia University, USA\\
\email{shiliangsun@gmail.com}
}
%
%
\toctitle{}
\tocauthor{}
\maketitle

\begin{abstract}
Dirichlet processes (DP) are widely applied in Bayesian nonparametric modeling. However, in their basic form they do not directly integrate dependency information among data arising
from space and time. In this paper, we propose location dependent Dirichlet processes (LDDP) which incorporate nonparametric Gaussian
processes in the DP modeling framework to model such dependencies.  We develop the LDDP in the context of mixture modeling, and develop a mean field variational inference
algorithm for this mixture model. The effectiveness of the proposed
modeling framework is shown on an image segmentation task.
\end{abstract}

\section{Introduction}

For many practical problems, nonparametric models are often chosen over
alternatives to parametric models that use a fixed and finite number
of parameters~\cite{BNP:2010}.
 In contrast, Bayesian
nonparametric priors are defined on
infinite-dimensional parameter spaces~\cite{Orbanz10BNM}, but fitting such models to data allows for an adaptive model complexity to be learned in the posterior distribution. In theory, this mitigates the underfitting and
overfitting problems faced by model selection for parametric models~\cite{Teh10DP}.

Dirichlet processes (DPs)~\cite{Ferguson73DP} are a standard Bayesian nonparametric prior for modeling data, typically via mixtures of simpler distributions. In this scenario, each draw
of a DP gives a discrete distribution on an infinite parameter space that can be used to cluster data into a varying number of groups.
While DPs have this flexibility as prior models for generating and modeling data, common additional data-specific markers such as time and space are often not incorporated in the mixture modeling formulation, and are simply ignored~\cite{Sun16UT,Zhu16BID}. However, for many problems this additional information can be an important part of data clustering. For example, in text models articles nearby
in time may be more likely to be clustered together by topics, and in image
segmentation, neighboring pixels are more likely to fall in the same category.
DP-based mixture models can often be improved by incorporating such information.

Such dependencies among the data are addressed in the literature through dependent
Dirichlet processes and their
generalizations~\cite{MacE99DNP,Griffin06Ob,Duan07Biom,Sudderth08SS,Griffin11OU}.
For example, the distance dependent Chinese restaurant process (ddCRPs) of Blei and Frazier~\cite{Blei11ddcrp} is a
clustering framework that uses a distance function between locations attached to data points to encourage clustering by their ``proximity.'' In the generative definition, it first
partitions data points by sequentially creating a linked network between observations, rather than by assigning data to clusters. The cluster assignments are then obtained as a by-product of the partition of the data according to the cliques in the ddCRP network.

While shown to be useful for spatial modeling \cite{Ghosh:2011}, the non-exchangeability of the ddCRP means that a mixing measure for such a process cannot be found along the lines of the stick-breaking construction for the DP. Therefore, there exists no distribution that makes all observations conditionally independent. As a result, the order of the data crucially matters for the ddCRP, which is often arbitrary and leads to local optimal issues. Since variational inference is a significant challenge as a result, Gibbs sampling was used for posterior inference of the ddCRP, which is computationally demanding and difficult to scale. Related exchangeable dependent random processes
based on beta process and probit stick-breaking processes have been recently
proposed, but for the mixed-membership model setting~\cite{Ren11KBP,Rod11NB}.

In this paper, we propose location dependent Dirichlet processes
(LDDPs) as a general dependent Dirichlet process modeling framework. Since a mixing measure for the ddCRP does not exist, our motivation is to define such a mixing measure for a model that achieves the same end goal, but is not equivalent to the ddCRP. To this end, we adapt ideas from the discrete infinite logistic normal (DILN) model \cite{Paisley12DILN} by combining Gaussian processes (GP) with Dirichlet processes. However, whereas DILN is a mixed-membership model that uses a single GP across latent cluster locations, the LDDP is a mixture model in which cluster-specific GP's interact directly with the data to capture distance dependencies. The direct definition of the LDDP mixing measure immediately allows for a variational inference algorithm to be derived. While the LDDP framework is general, we apply it to the Gaussian mixture model for image segmentation.


\section{LDDPs and an Inference Algorithm}
We first review the connection between Dirichlet and gamma processes and define the generative process of the location dependent Dirichlet process (LDDP). We then derive mean field variational inference with
the general LDDP and discuss a
proposed model for Gaussian data. We note that the term ``location'' refers to any auxiliary
information connected to the primary data, such as time or space.

\subsection{DPs and the Gamma Process}
The DP is a prior widely used for Bayesian nonparametric mixture modeling.
A draw $G$ from a DP with concentration parameter $\alpha_0$ and
base distribution $G_0$, written as $G\sim DP(\alpha_0 G_0)$, is a discrete probability distribution on the support of $G_0$. Suppose
\begin{equation}
v_i \stackrel{iid}{\sim} Beta(1, \alpha_0), \quad \theta_i^*
\stackrel{iid}{\sim} G_0 \;,
\end{equation}
and define $\pi_i= v_i \prod_{j=1}^{i-1}(1-v_j)$. A way of constructing the infinite distribution $G$ is \cite{Sethuraman:ss94}
\begin{equation}
G=\sum_{i=1}^\infty \pi_i \delta_{\theta_i^*} \;.
\end{equation}
With the DP mixture, data are generated independently as,
\begin{equation}
\theta_n|G   \stackrel{iid}{\sim} G,\quad x_n|\theta_n \sim p(x|\theta_n) \;.
\end{equation}
A partition of the data is naturally formed according to the repeating of
atoms $\{\theta_i^*\}$ among the parameters $\{\theta_n\}$ that are used by the data.

It is well-known that the DP can be equivalently represented as an
infinite limit of a finite mixture
model, and through normalized gamma measures~\cite{Ferguson73DP,Ishwaran:2002a}. In this case, suppose there are $K_0$
components in the finite mixture model, and
$$ \theta_i^* \stackrel{iid}{\sim} G_0, \quad z_i \stackrel{iid}{\sim} Gamma(\textstyle\frac{\alpha_0}{K_0},1) \quad {G}_{K_0} = \sum_{i=1}^{K_0} \frac{z_i}{\sum_{j=1}^{K_0}z_j}
\delta_{\theta_i^*}.$$
Then as $K_0\rightarrow\infty$, ${G}_{\infty}\sim DP(\alpha_0 G_0)$. For computational convenience, we can form an accurate approximation to the DP by using $G_{K_0}$ with a large value of $K_0$ \cite{Ishwaran:2002a}.

\subsection{Location Dependent Dirichlet Processes}
We extend the DP to the LDDP by associating with the atom $\theta_i^*$ of each cluster a Gaussian process $$\qquad {f}_i(\ell)\sim \mathcal{GP}(0,
k(\ell,\ell^\prime)), \quad i=1,2,\dots$$ on a particular space of interest, $\ell \in \Omega$. For example, the $\ell$ indicates geographic location or is a time stamp. We note that the Gaussian process of each cluster is defined on, e.g., all time or space. Our goal is to allow the associated location $\ell_n$ for observation $x_n$ to \textit{increase} the probability of using cluster parameter $\theta_i^*$ when $f_i(\ell_n) > 0$, and \textit{decrease} that probability when $f_i(\ell_n) < 0$. The kernel of the Gaussian process $k(\cdot,\cdot)$ ensures that each cluster marks off contiguous regions in space or time. For example, we use the common Gaussian kernel in our experiments,
\begin{equation}\label{eqn.rbf}
 k(\ell,\ell') = \sigma_f^2 \exp\left[-\|\ell-\ell'\|^2/\sigma_{\ell}^2\right].
\end{equation}
We observe that GPs generated with such a kernel will be continuous and are flexible enough to be positive or negative in various regions of space \cite{Rasmussen:2006}, which provides more modeling capacity than the ddCRP.

The LDDP uses these GPs in combination with a gamma process representation of the DP to generate an \textit{observation-specific} distribution on clusters. Employing the finite-$K_0$ approximation to the DP above, we again let $G_0$ be the base distribution for
$\{\theta_i^*\}$. Suppose $c_n$ is a discrete latent variable which
indicates the atom assigned to $x_n$, so that $\theta_n = \theta_{c_n}^*$. We first generate
$$\theta_i^* \stackrel{iid}{\sim} G_0, \quad z_i \stackrel{iid}{\sim} Gamma(\textstyle\frac{\alpha_0}{K_0},1)$$
exactly as before. Then, for observation $n$ our construction of the LDDP distribution on the clusters is
\begin{equation}
P(c_n=k|\boldsymbol{z},\boldsymbol{f},\ell_n) ~ \propto ~ z_k e^{f_k(\ell_n)},\label{eqn_cn}
\end{equation}
for each observation $n=1,\dots,N$. This is a trade-off between how prevalent cluster $k$ is globally\,---\,$z_k$\,---\,and how appropriate cluster $k$ is for the $n$th observation\,---\,$e^{f_k(\ell_n)}$.  Using the previous notation, this can also be written
\begin{equation}\label{eqn.Gn}
G_n=\sum_{i=1}^{K_0} \frac{z_i e^{f_i(\ell_n)}}{\sum_{j=1}^{K_0} z_j e^{f_j(\ell_n)}} \delta_{\theta_i^*},
\end{equation}
from which we generate data
\begin{equation}
 \theta_n|G_n \sim G_n, \quad x_n|\theta_n \sim p(x|\theta_{n}) .
\end{equation}

Since the atoms $\theta_i^*$ are shared among each distribution $G_n$, a
partition of the data is formed according to the values of the indicator
variables $c_1,\dots,c_N$. However, as is clear from Eq.\ (\ref{eqn.Gn}), each observation $x_n$ does not use these atoms i.i.d.\ as in the standard DP. Instead, the Gaussian processes encourages those $x_n$ that have auxiliary information $\ell_n$ in positive regions of the Gaussian processes to cluster together. These will tend to cluster $x_n$ with $\ell_n$ that are close (e.g., in time or space). We note that we do not define a generative model for these $\ell_n$, but only $x_n | \ell_n$. In posterior inference, clustering will be a trade-off between how similar two $\ell_n$ are according to the Gaussian process, and how similar two $x_n$ are according to the data distribution $p(x|\theta)$.

%
%

\subsection{Mean-Field Variational Inference}
We let the data-generating distribution $p(x|\theta)$ be generic for the moment and discuss a variational inference algorithm for the LDDP in general.
Given $N$ observations with corresponding location variables $\{(x_n,\ell_n)\}$, the joint distribution of
the model variables and data factorizes as
\begin{equation}
\textstyle p(\bs{x},\bs{c},\bs{z},\bs{f},\bs{\theta}|\bs{\ell}) = p(\bs{z},\bs{f},\bs{\theta})\prod_n p(x_n|\theta_{c_n})p(c_n|\bs{z},\bs{f},\ell_n).
\end{equation}
We derive a variational inference algorithm for the sets of variables $\bs{z}$, $\bs{f}$ and $\bs{c}$, which occur in all potential LDDP models. We recall that with mean-field variational inference \cite{Bishop06prml,Blei14BCCR}, we define a factorized approximation to the posterior distribution, $$\textstyle p(\bs{c},\bs{z},\bs{f},\bs{\theta}|\bs{\ell},\bs{x}) \approx \big[\prod_n q(c_n)\big]\big[\prod_k q(z_k)q(f_k)q(\theta_k)\big].$$ After choosing specific distributions for each $q$,\footnote{$q(\theta_k)$ is problem-specific, so we ignore it here.} we then tune the parameters of these distributions to maximize the variational objective function
$$\mathcal{L} = \mathbb{E}_q[\ln p(\bs{x},\bs{c},\bs{z},\bs{f},\bs{\theta}|\bs{\ell})] - \mathbb{E}_q[\ln q].$$
Coordinate ascent is usually adopted to optimize the objective by cycling through optimizing each $q$ within each iteration. For the LDDP model, we choose,
$$q(z_k) = Gam(a_k,b_k),~ q(c_n) = Mult(\phi_n),~ q(f_k) = \delta_{f_k}.$$
The last choice is out of convenience, since a distribution of $f_k$ (an $N$ dimensional vector) has computationally-intensive tractability issues. A delta $q$ distribution amounts to a point estimate of the variable in the objective function $\mathcal{L}$.

The joint distribution $p(\bs{x},\bs{c},\bs{z},\bs{f},\bs{\theta}|\bs{\ell})$ presents further difficulties, which can be seen by expanding it as
\begin{equation}
\prod_{n=1}^{N}\prod_{i=1}^{K_0}\Big( p(x_n|c_n)\frac{z_i
e^{f_i(\ell_n)}}{\sum_{j=1}^{K_0} z_j e^{f_j(\ell_n)}}
\Big)^{1(c_n=i)}\Big[\prod_{i=1}^{K_0} z_i^{\frac{\alpha_0}{K_0}-1}e^{-z_i}\Big]
\Big[ \prod_{i=1}^{K_0} e^{-\frac{1}{2}{f}_i^\top K^{-1}
{f}_i}\Big]. \label{eqn_zfc}
\end{equation}
The normalization of $z_i e^{f_i}$ makes directly calculating $\mathcal{L}$ intractable, since \linebreak $\mathbb{E}_q[-\ln \sum z_j e^{f_j(\ell_n)}]$ is not in closed form when integrating over each $z_j$.
We therefore use a lower bound of this term found useful in similar situations, e.g., \cite{Paisley12DILN}. Introducing an auxiliary parameter $\xi_n > 0$, by a simple first order Taylor expansion of the convex function $-\ln(\cdot)$ we have
\begin{equation}
-\ln \sum_j z_j e^{f_j(\ell_n)}\geq -\ln\xi_n -\frac{\sum_j z_j
e^{f_j(\ell_n)}-\xi_n}{\xi_n}.
\end{equation}
Therefore, in the joint likelihood we replace
\begin{equation}
\frac{1}{\sum_j z_j e^{f_j(\ell_n)}} ~\geq~ \frac{1}{\xi_n}
~e^{-\xi_n^{-1}\sum_j z_j e^{f_j(\ell_n)}}.
\end{equation}
Differentiating the new objective with respect to  $\xi_n$ and setting to zero, we see that the lower bound is tightest at
\begin{equation}
\textstyle\xi_n =\sum_j \mathbb{E}_q [z_j] e^{f_j(\ell_n)} . \label{eqnxin}
\end{equation}
Thus, $\xi_n$ becomes a new parameter in the model that is set to this value at the end of each iteration. In this and all following equations, the expectations are calculated using the most recent parameters of the relevant $q$ distribution.

For the remaining $q$ distributions, following the steps in \cite{Bishop06prml} for $q(c_n)$ and $q(z_k)$, the multinomial distribution $q(c_n)$ can be updated at each iteration by setting its discrete distribution parameter
\begin{equation}\label{eqnqcn}
 \phi_n(k) \propto \exp\{\mathbb{E}_q[\ln p(x_n|\theta_k)] + \mathbb{E}_q[\ln z_k] + f_k(\ell_n)\}.
\end{equation}
The first expectation is problem-specific and depends on the data $x_n$ and the distributions chosen for modeling it and $\theta_k$.

The parameters for the gamma distribution $q(z_k)$ can be updated by setting them to
\begin{equation}
a_k = \frac{a}{K_0} +\sum_{n=1}^N \mathbb{E}_q [1(c_n=k)], ~~
b_k = 1 + \sum_{n=1}^N\frac{e^{f_k(\ell_n)}}{\xi_n}.\label{eqnqz}
\end{equation}

To update each Gaussian process $f_k$ at the $N$ locations, we use gradient ascent. The gradient $\nabla_{{f}_k}\mathcal{\widetilde{L}}$, where the tilde indicates the lower bound approximation to $\mathcal{L}$, is
\begin{equation}\label{eqn.f}
\nabla_{{f}_k}
\widetilde{\mathcal{L}}=\Big[\mathbb{E}_q[1(c_n=k)]-\frac{1}{\xi_n}\mathbb{E}_q
[z_k]e^{f_k(\ell_n)} \Big]_n - \mathbf{K}^{-1}{f}_k .
\end{equation}
We take a gradient step using $\mathbf{K}$ as a convenient preconditioner (discussed more below)
\begin{equation}\label{eqn.f2}
 {f}_k \leftarrow {f}_k + \rho \mathbf{K} \nabla_{{f}_k}\widetilde{\mathcal{L}}.
\end{equation}

\subsection{Computational Considerations} When the number of observations $N$ is large, the $N\times N$ kernel $\mathbf{K}$ can be massive. Not only is the inverse not feasible in this situation, but calculating $\mathbf{K}$ itself results in memory issues. We use a simple approach based on the Nystr\"{o}m method to address this issue \cite{Williams01nys,Kumar12SM}.

Specifically, let $\bs{\ell}^*$ be a set of $N_2 \ll N$ locations in the same space as $\bs{\ell}$. These $N_2$ locations can be different from those in the data set, and should be spread out in the space. For example, in an image these might be $N_2$ evenly spaced grid points. Then let $\mathbf{K}^*$ be the kernel restricted to these $N_2$ locations, and $\mathbf{K}^{**}$ the kernel between $\bs{\ell}^*$ and $\bs{\ell}$, so that $$\mathbf{K}^*_{i,j} = k(\ell_i^*,\ell_j^*),\quad \mathbf{K}^{**}_{i,j} = k(\ell_i^*,\ell_j).$$ Then it is well-known that for appropriately chosen $\bs{\ell}^*$, an accurate approximation to the $N\times N$ kernel $\textbf{K}$ is
\begin{equation}\label{eqn.Kapprox}
 \textbf{K} ~\approx ~ (\textbf{K}^{**})^\top (\textbf{K}^*)^{-1}\textbf{K}^{**}
\end{equation}
As a result, when updating each ${f}_k$ as in Eq.\ (\ref{eqn.f2}), we only need to work with $N_2\times N_2$ and $N_2\times N$ matrices. These matrices are much smaller and can be calculated in advance and stored for re-use, and so an $N\times N$ matrix never needs to be constructed. We also note that this approximation is being performed (in principle) after multiplying $\mathbf{K} \nabla_{{f}_k}\widetilde{\mathcal{L}}$, and so we do not need to approximate $\mathbf{K}^{-1}$.

\subsection{LDDP Mixtures of Gaussian Distributions}

\begin{wrapfigure}{r}{0.45\textwidth}
\vspace{-20pt}
\includegraphics[width=.45\textwidth]{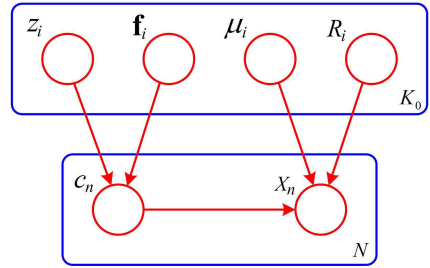}  
\caption[]{The graphical model for the LDDP mixture of Gaussian distributions. This model extends the GMM by including a Gaussian process $f_i$ in the cluster assignment prior, which encourages region-based clustering.}
\label{figlddpgm}
\vspace{-25pt}
\end{wrapfigure}

We apply the LDDP prior to mixture models for which the data are Gaussian. In this case, $\theta_i^*=\{\mu_i, R_i\}$ where  $\mu_i$ and
$R_i$ are the mean and inverse covariance of a multivariate Gaussian distribution. We also specify the priors for  $\mu_i$ and $R_i$ as normal and Wishart
distributions as $\mu_i \sim \mathcal{N}(\mu_0, R_0^{-1}), ~ R_i \sim \mathcal{W}(W_0, \nu_0) .$
The graphical model for the LDDP mixture of Gaussian distributions
is given in Figure~\ref{figlddpgm}, where the hyperparameters are
not shown. Inference details for this model are standard, and thus omitted here.

\subsubsection{Some Applications:} An example we consider in our experiments is image segmentation. In this setting, each $x_n$ could be the 3-D RGB vector of pixel $n$. The location $\ell_n$ would then be the 2-D coordinates of this pixel in the image. Each cluster would consist of a 3-D Gaussian distribution on RGB, which would cluster similar colors, and a Gaussian process that would indicate which regions of the image this cluster would be more likely to be active. This GP would be intended to improve the segmentation over a direct Gaussian mixture model (GMM).
Another example for future consideration would be audio segmentation. The setup would be almost identical to image segmentation, however in this case $x_n$ could be a short-time frequency content vector, such as an MFCC, and $\ell_n$ would be the time stamp within the audio. A third example could capture geographic information in $\ell_n$, and a feature vector $x_n$ for the person or business with index $n$ having the location $\ell_n$.


\section{Discussion}
The LDDP is a type of dependent DP where Gaussian
processes are involved to adapt the generating probabilities of the
atoms in a nonparametric manner. In addition to ddCRPs, our model is also
related to but still different from several other dependent DPs which
we briefly discuss here.

The kernel stick-breaking process~\cite{Dunson08KSBP} is constructed by introducing a countable sequence of mutually independent random
variables
\begin{equation}
\{\Gamma_h, V_h, G_h^*, h=1, \ldots, \infty\} ,
\end{equation}
where $\Gamma_h\sim H$ is a location, $V_h\sim Beta(a_h, b_h)$, and
$G_h^*\sim \mathcal{Q}$ is a probability measure. Then, the process
is defined as follows:
\begin{eqnarray}
&&G_{x_n}=\sum_{h=1}^\infty U(x_n, V_h,
\Gamma_h)\prod_{i<h}\{1-U(x_n,
V_i, \Gamma_i)\}G_h^*, \nonumber\\ &&U(x_n, V_h, \Gamma_h)
= V_h k(x_n, \Gamma_h),
\end{eqnarray} where
$k(\cdot,\cdot)$ is a kernel
function. The kernel stick-breaking process accommodates dependency
since for close $x_n$ and $x_{n^\prime}$, the $G_{x_n}$ and
$G_{x_{n^\prime}}$ will assign similar probabilities to the elements
of $\{G_h^*\}_{h=1}^\infty$. By inspection, we find that the table
assignments induced from the kernel stick-breaking process are
generally not exchangeable but marginally invariant. This process was later
extended to hierarchical kernel stick-breaking
process~\cite{An08HKSBP} for multi-task learning.

Foti and Williamson~\cite{Foti12SS} introduced a large class of dependent nonparametric
processes which are also similar to LDDPs, but uses parametric kernels
to weight dependency. Foti et al.~\cite{Foti13UP} presented a general
construction for dependent random measures based on thinning Poisson
processes, which can be used as priors for  a large class of
nonparametric Bayesian models. In contrast to our LDDP, the
proportion variable of the thinned completely random measures comes
from the global measure, and the rate measures involve parametric
formulations.

\section{Experiments}
\begin{figure}[!hp]
\begin{center}
 \includegraphics[width=.24\columnwidth]{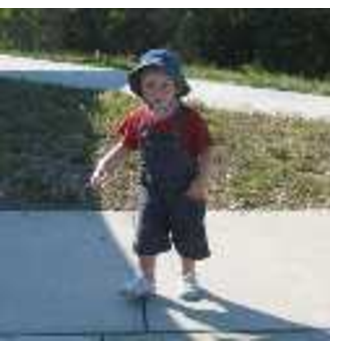}
 \includegraphics[width=.24\columnwidth]{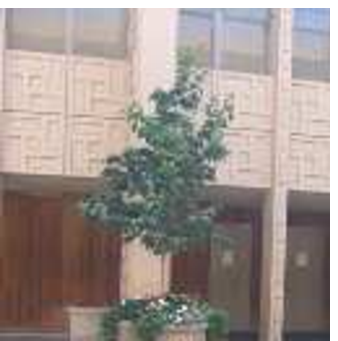}
 \includegraphics[width=.24\columnwidth]{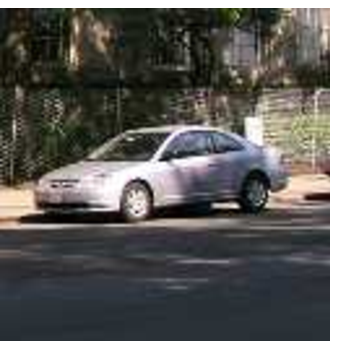}
 \includegraphics[width=.24\columnwidth]{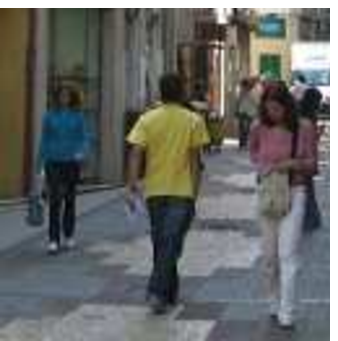}\\  
 \includegraphics[width= .24\columnwidth]{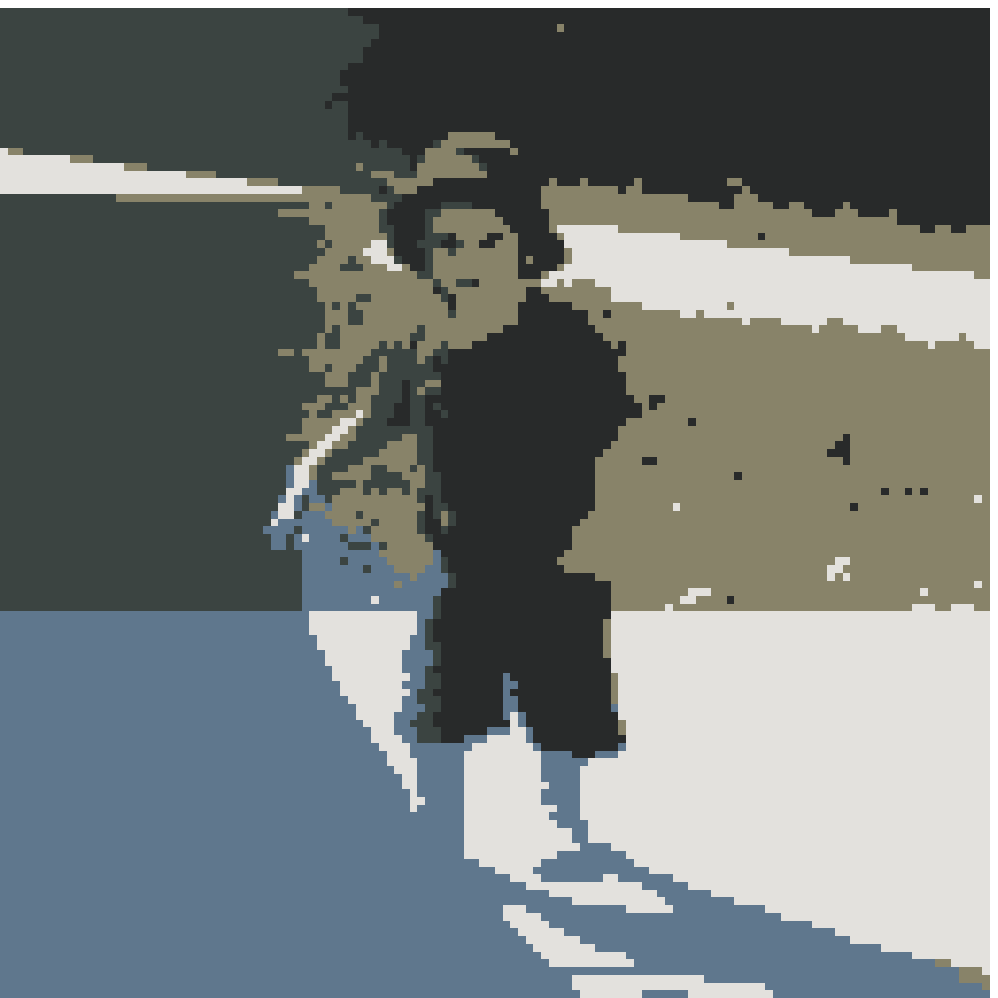}
 \includegraphics[width= .24\columnwidth]{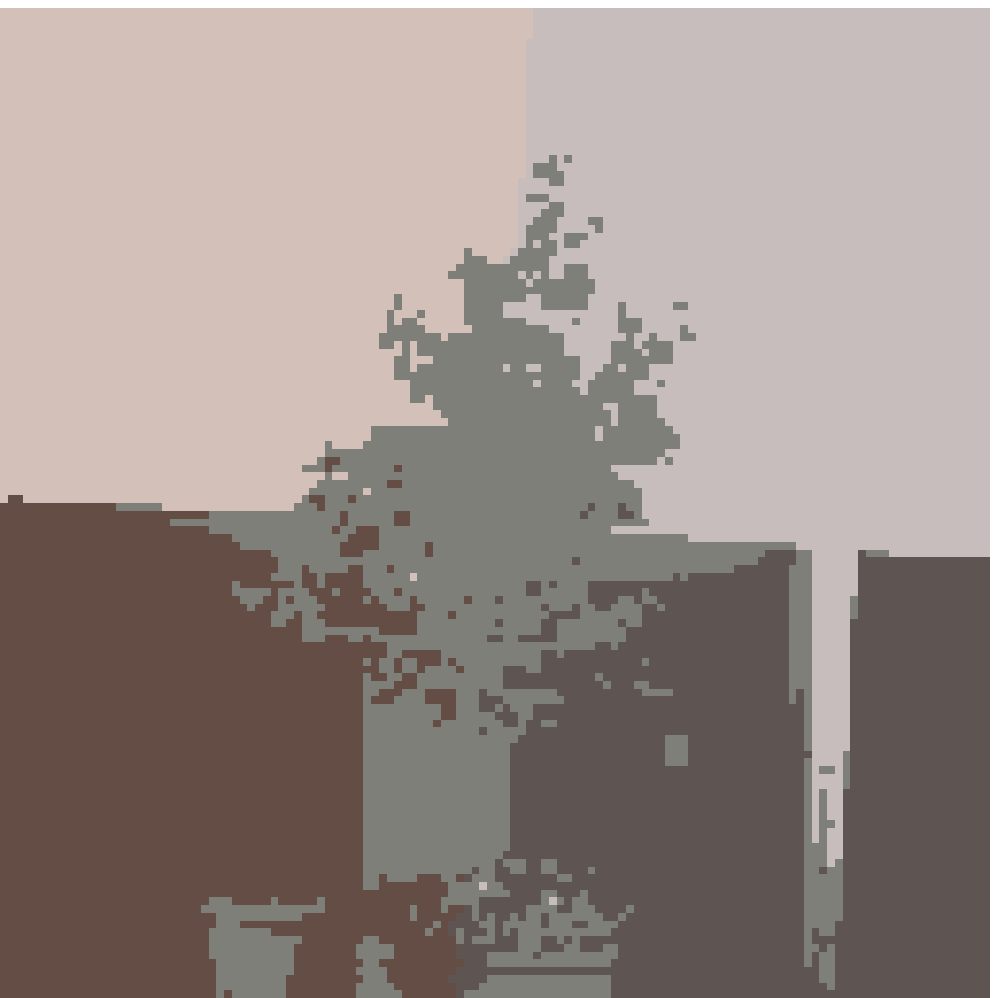}
 \includegraphics[width= .24\columnwidth]{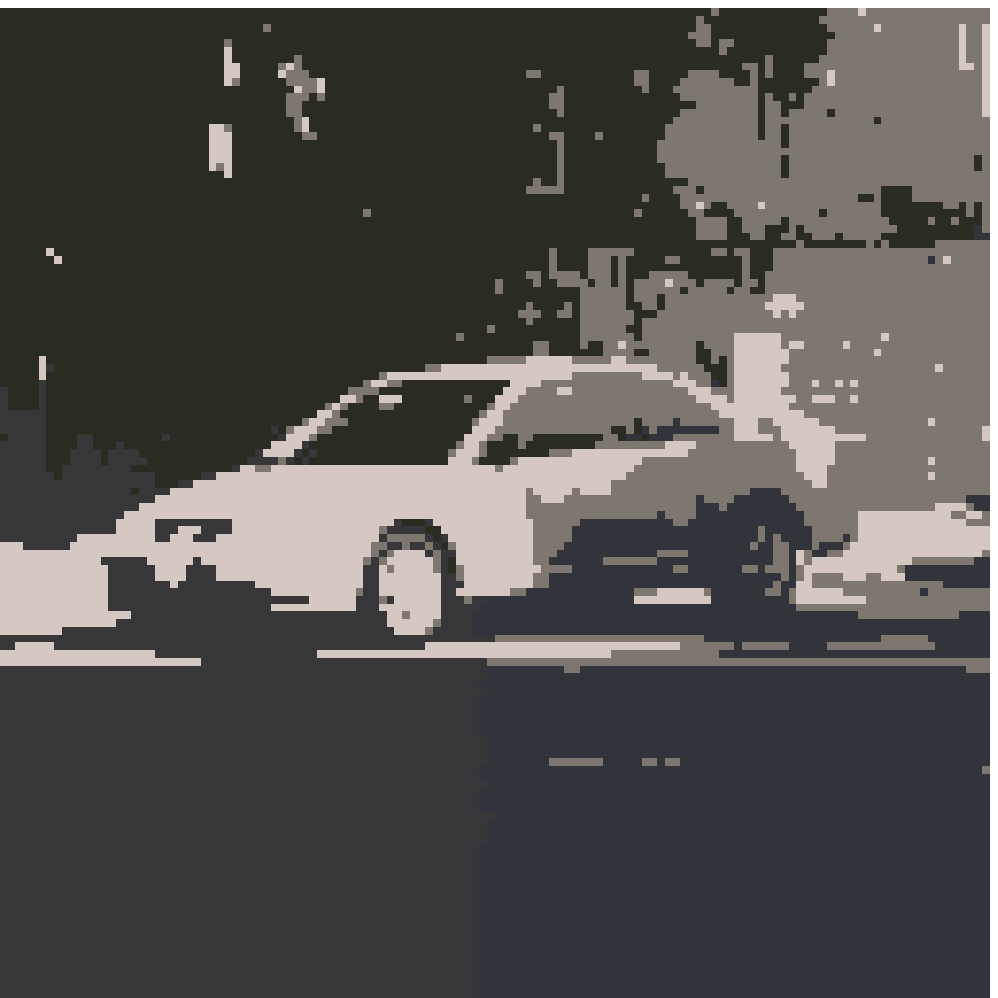}
 \includegraphics[width= .24\columnwidth]{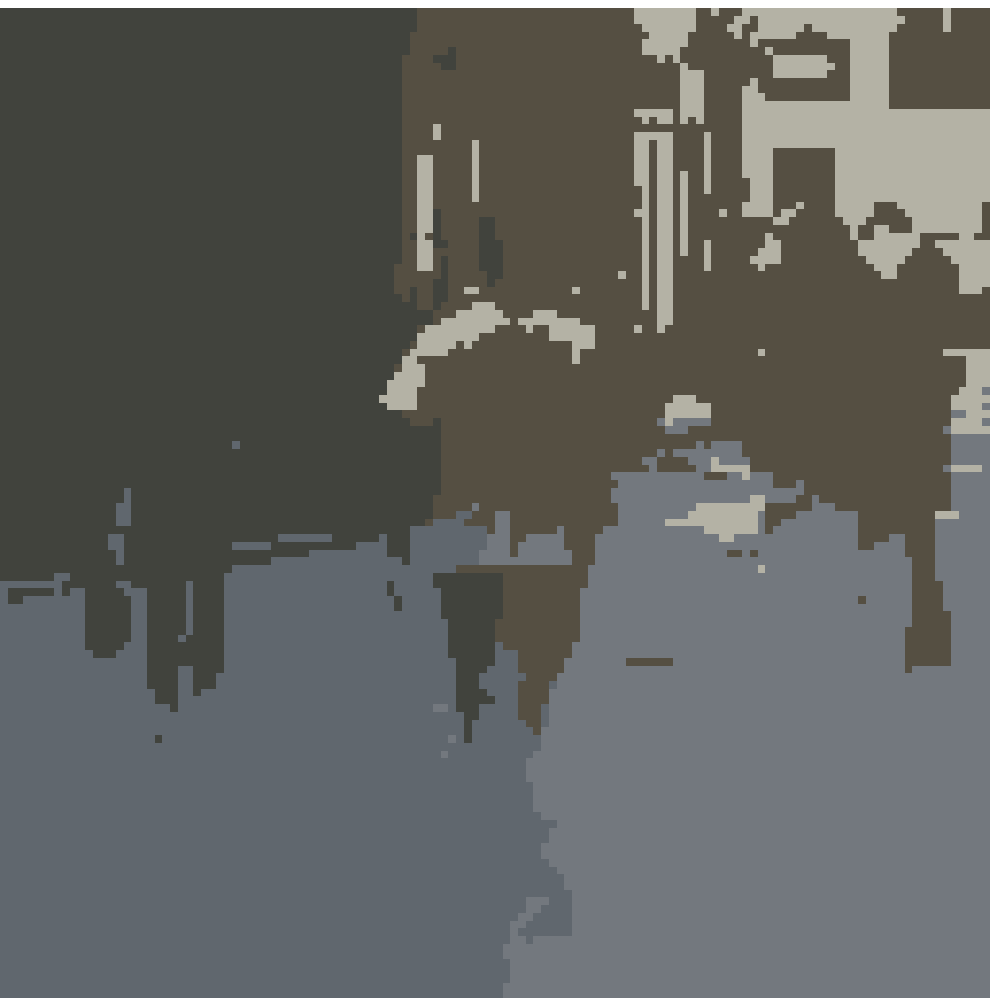}\\

  \includegraphics[width= .24\columnwidth]{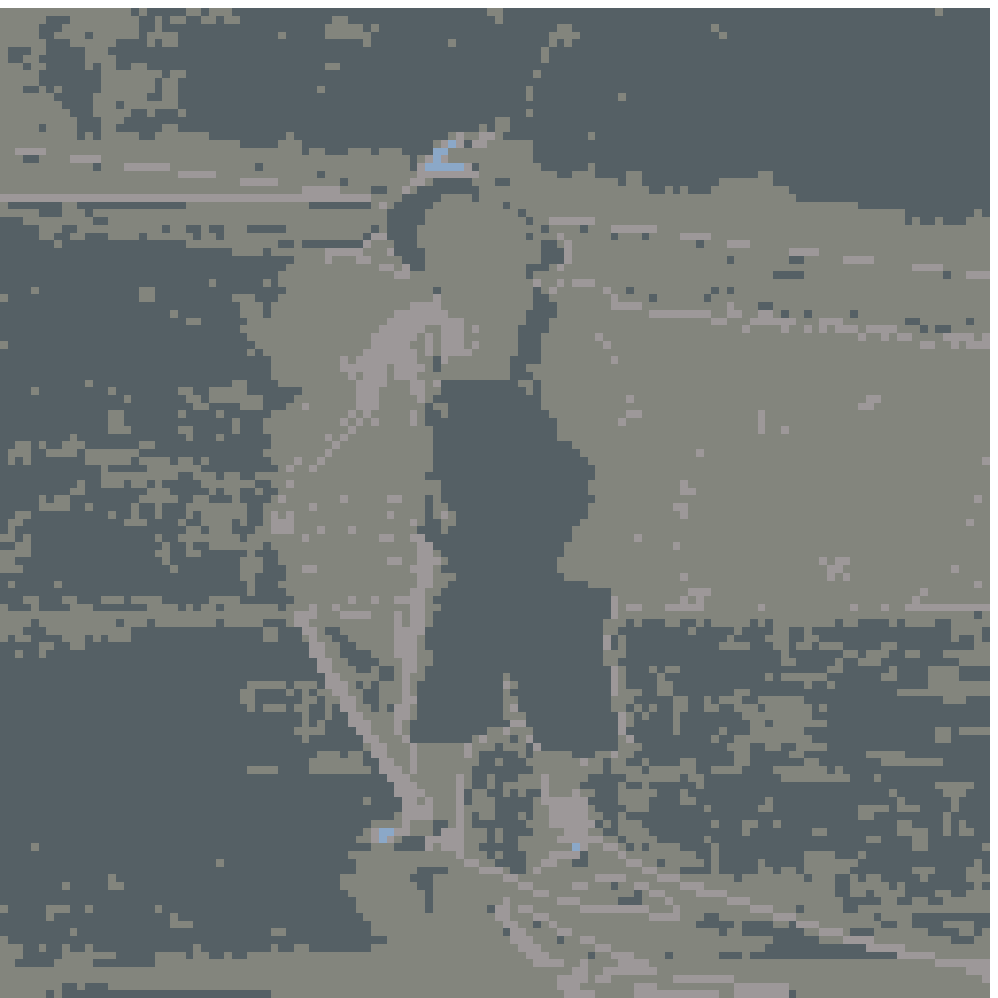}
 \includegraphics[width= .24\columnwidth]{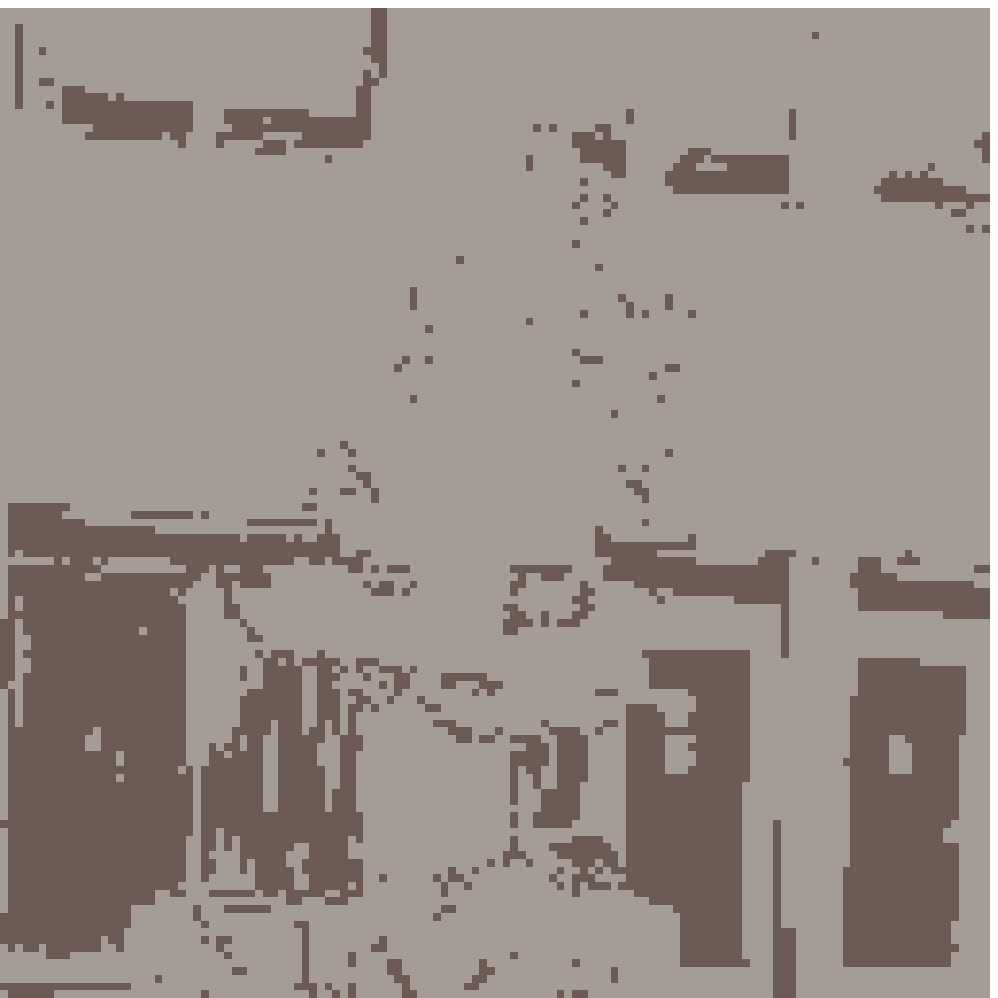}
 \includegraphics[width= .24\columnwidth]{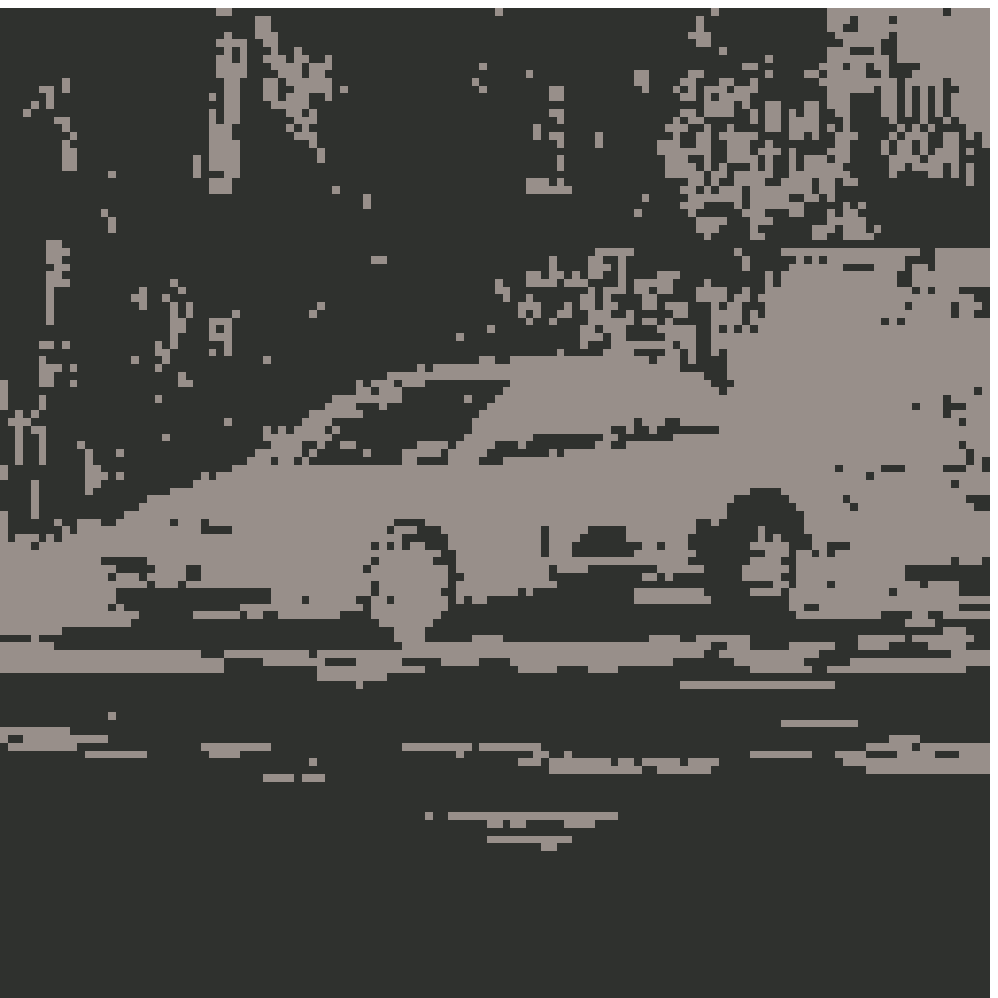}
 \includegraphics[width= .24\columnwidth]{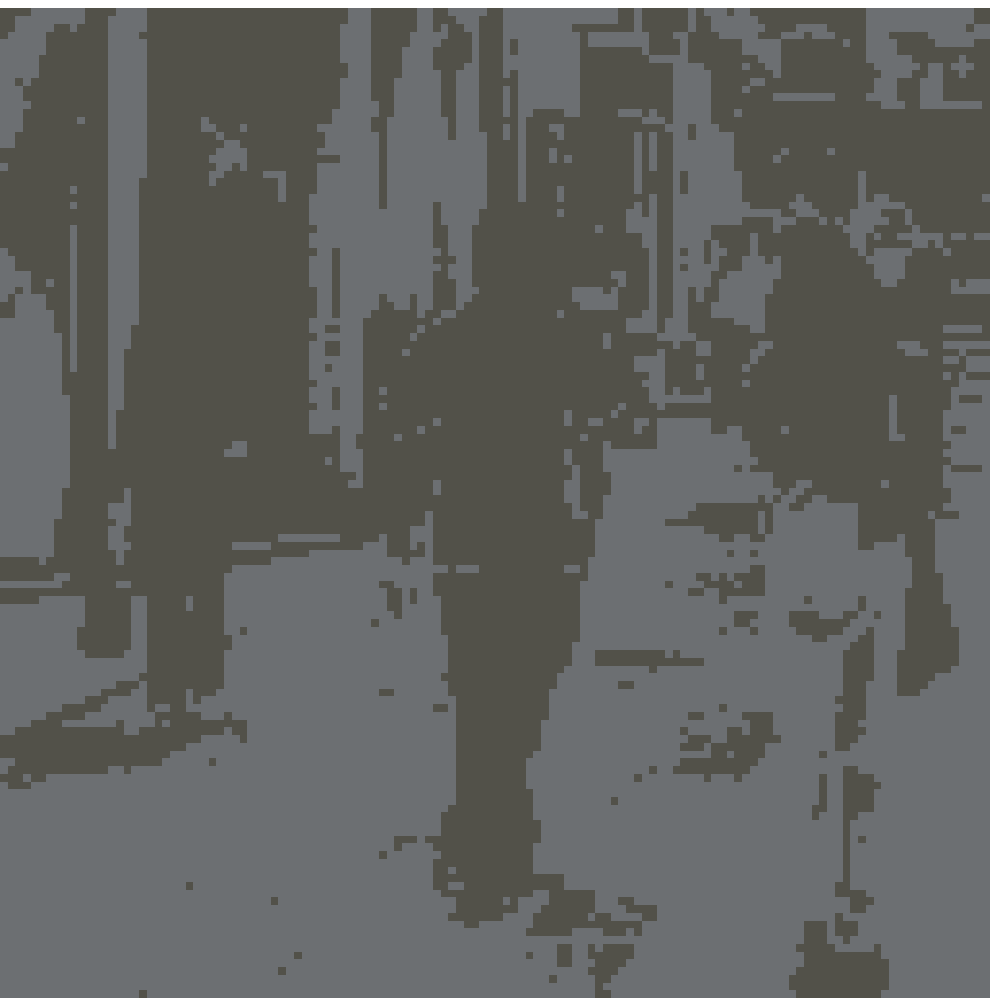}\\

 \includegraphics[width= .24\columnwidth]{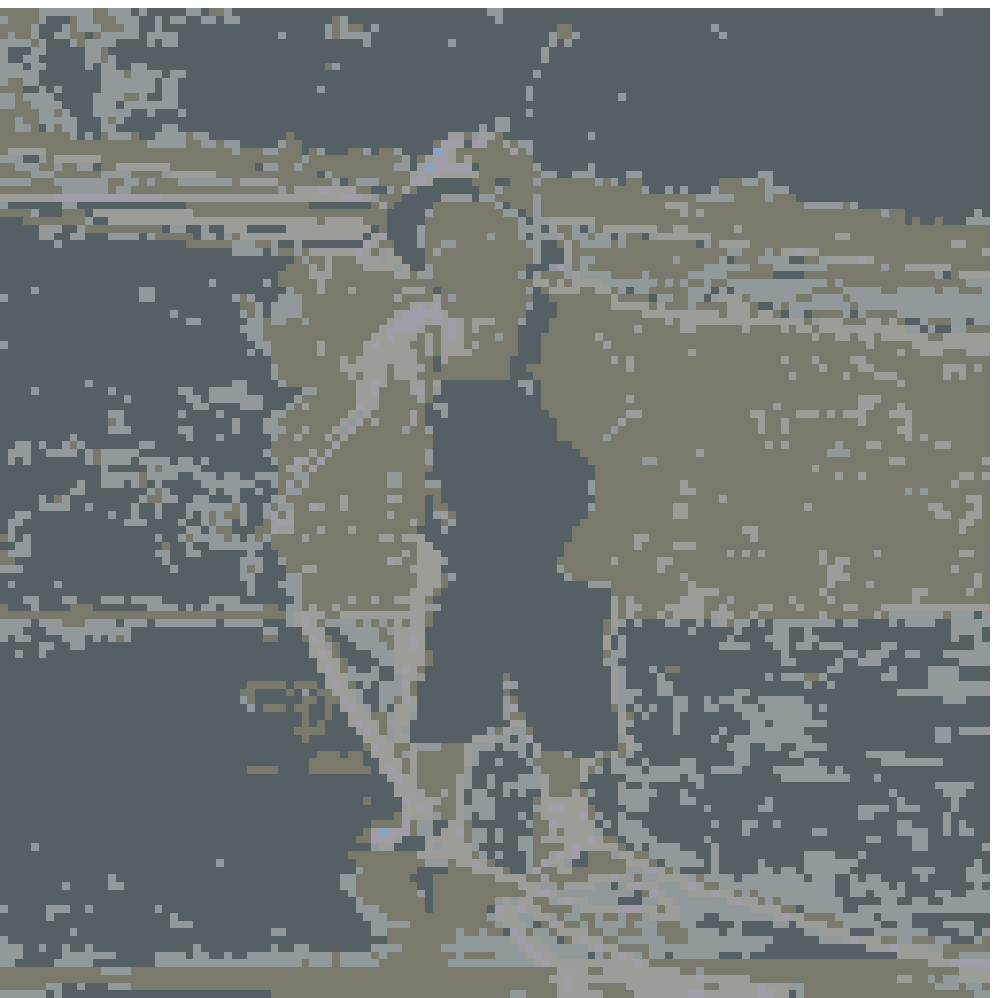}
 \includegraphics[width= .24\columnwidth]{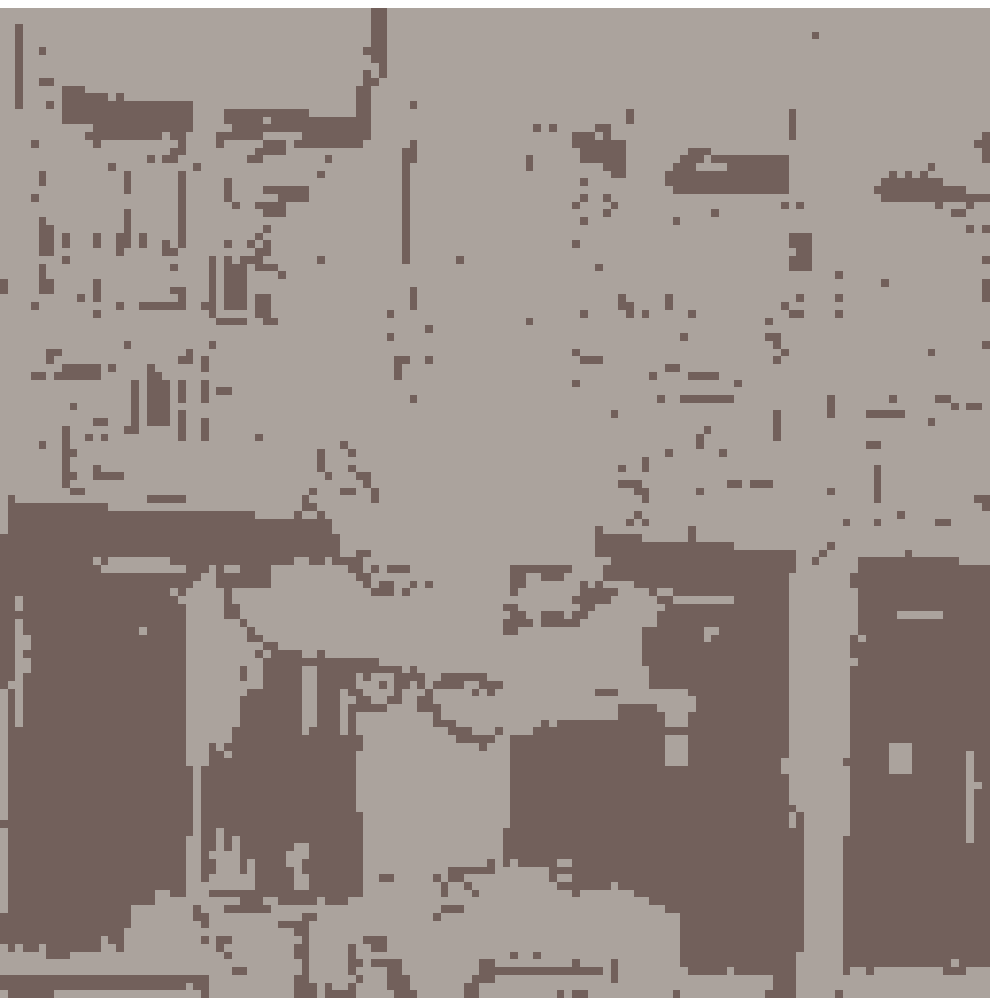}
 \includegraphics[width= .24\columnwidth]{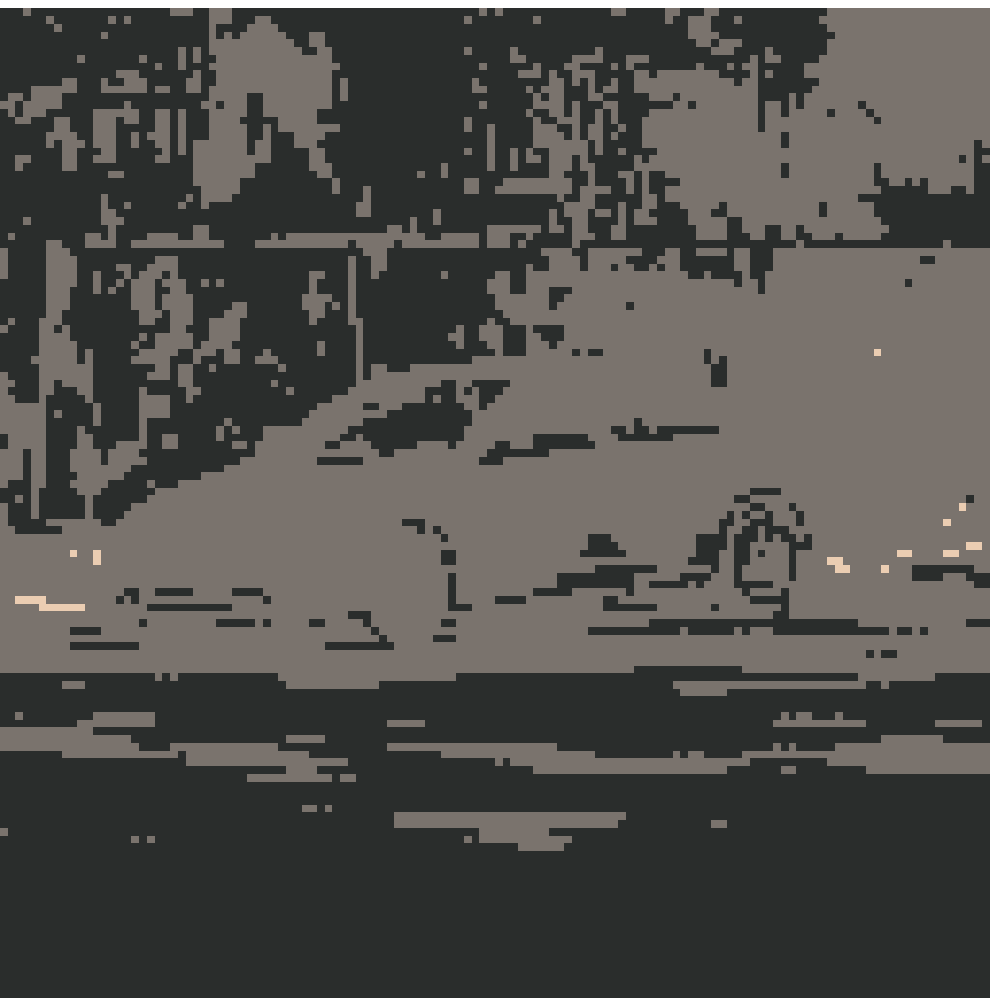}
 \includegraphics[width= .24\columnwidth]{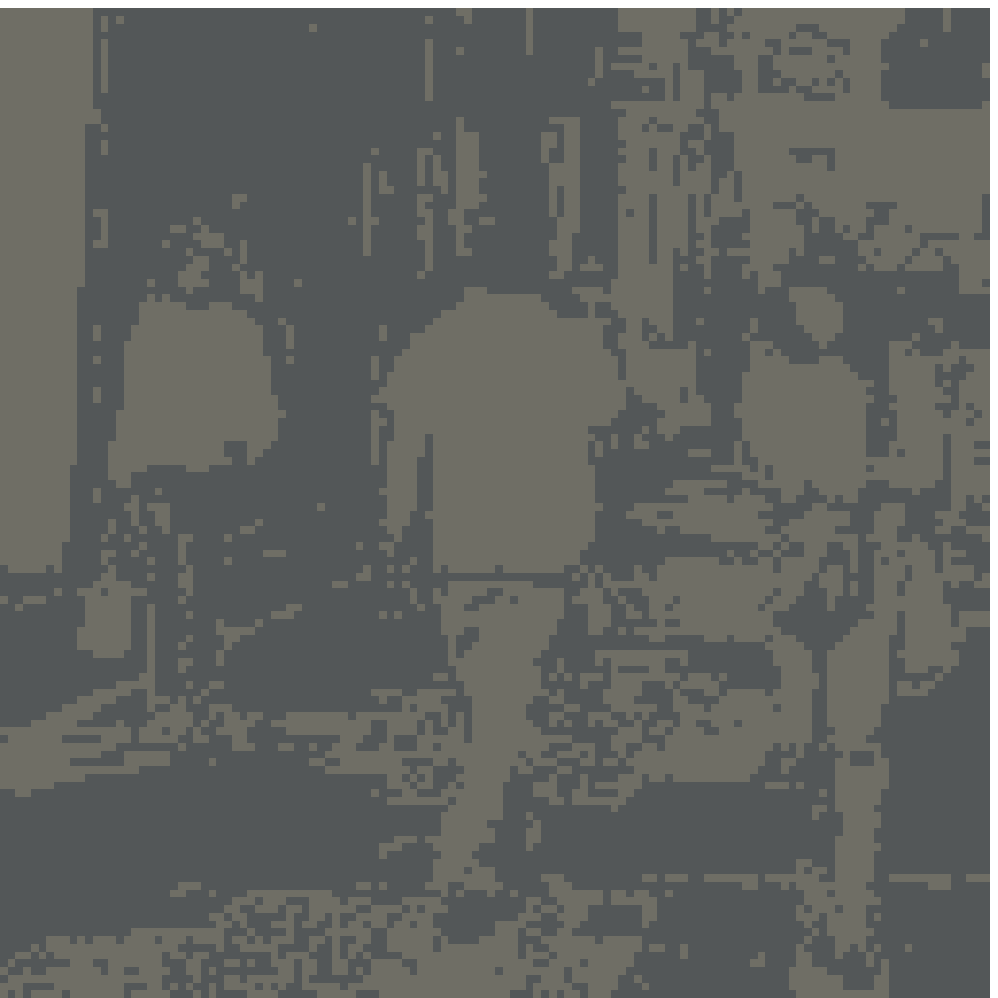}\\  

 \includegraphics[width=.24\columnwidth]{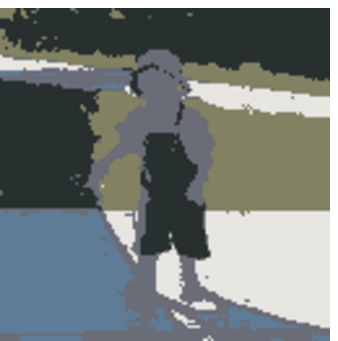}
 \includegraphics[width=.24\columnwidth]{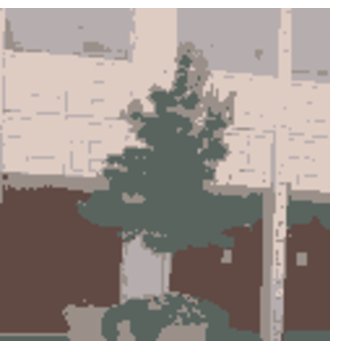}
 \includegraphics[width=.24\columnwidth]{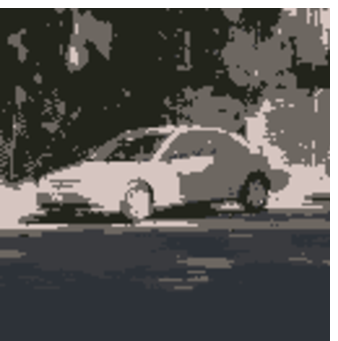}
 \includegraphics[width=.24\columnwidth]{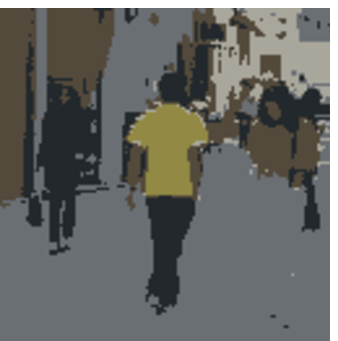}\\
  \includegraphics[width=.24\columnwidth]{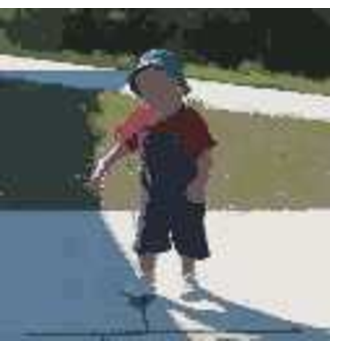}
  \includegraphics[width=.24\columnwidth]{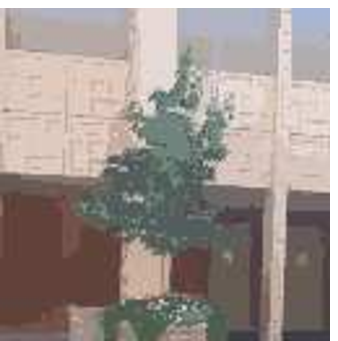}
  \includegraphics[width=.24\columnwidth]{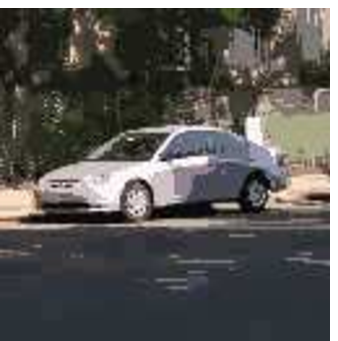}
  \includegraphics[width=.24\columnwidth]{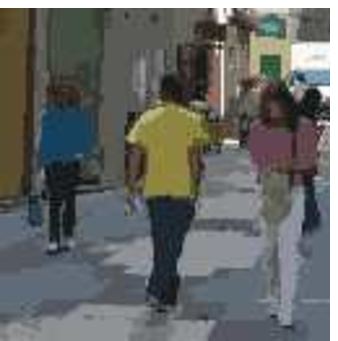}
\end{center}
\caption[]{Original images (first row) and segmentation results (other rows). The second row is obtained
  by $K$-means with RGB and pixel locations.
 The third and fourth rows are DPYP and HPY, respectively.
 The fifth and sixth rows are the proposed LDDP using $K_0=5$ and $K_0 = 100$.}
\label{fig4Images}
\end{figure}

\subsection{Setup}
We define the kernel function to be the radial basis function (RBF) in Eq.\ (\ref{eqn.rbf}) with settings for $\sigma_f$ and $\sigma_{\ell}$ described below.
For hyperparameters, we set $\alpha_0 = 1$ and ${\mu}_0$ and ${R}_0$ are set to the empirical mean and inverse covariance of the training data $\bs{x}$. The Wishart parameter $\nu_0$ is set to the dimensionality $d$
of the training data, with $d=3$ for our problems. ${W}_0$ is set to ${R}/d$ such that
the mean of ${R}_i$ under the Wishart distribution is ${R}$. For the $q$ distributions, we initialize each $f_i$ to be the zero vector, and set each $a_i = b_i = 1$. We define $q(\theta_i) = q(\mu_i)q(R_i)$ to be normal-Wishart and initialize them to be equal to the prior, with the important exception that the mean of $q(\mu_i)$ is initialized using $K$-means.

We apply the Gaussian LDDP to a segmentation problem of natural scene images. The size of the images we used is $128\times 128$, and thus the size of the kernel matrix
of the Gaussian process is $16384 \times 16384$. We use the Nystr\"{o}m method here with approximately $5\%$ of these locations evenly spaced in the image.

For comparison, we use the $K$-means clustering algorithm as a baseline for performance
comparison. We also compare our method with normalized cut spectral clustering \cite{Shi00NC}, dependent Pitman-Yor processes (DPYP) \cite{Sudderth08SS} and hierarchical Pitman-Yor (HPY) \cite{HPY}, as well as special cases of the LDDP such as the GMM. We run all algorithms for $1000$ iterations, which empirically were sufficient for convergence.

\subsection{Image Segmentation Results}
We show segmentation results using images of different scenes from the LabelMe data set~\cite{Uetz09LSOB} in Figure~\ref{fig4Images}. In our experiments, we consider a parametric version of the LDDP in which $K_0 = 5$ and a nonparametric approximation, where $K_0 = 100$. These two cases were differentiated by the posterior cluster usage, where all were used in the first case and only a subset used in the second.

\begin{wrapfigure}{r}{0.5\textwidth}\vspace{-20pt}
  \centering
  \includegraphics[width=.5\textwidth]{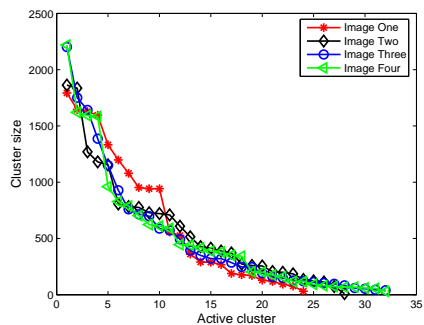}
  \caption{\#pixels assigned to active clusters}
  \label{figClustersize}\vspace{-10pt}
\end{wrapfigure}

We compare with $K$-means segmentation in which we use a 5-D vector, three for RGB and two for the pixel location in the image. These results are shown in the second row of Figure~\ref{fig4Images}, where five clusters are used (i.e., $K_0=5$). The dependent Pitman-Yor processes (DPYP) segmentation results for the  images are shown in the third row. The DPYP uses thresholded Gaussian processes to generate spatially coupled Pitman-Yor processes. The hyperparameters involved in the DPYP are set analogously to the  ones in \cite{Sudderth08SS}.  For the covariance functions involved in the DPYP, we use the distance-based squared exponential covariance, which has been shown to give good results \cite{Sudderth08SS}.
The DPYP includes the hierarchical Pitman-Yor (HPY) model as a special case when the Gaussian processes involved have identity covariance functions. The HPY mixture segmentation results for the  images are shown in the fourth row of Figure~\ref{fig4Images}.

For the LDDP mixture, when $\sigma_{\ell}=0.1$ and $\sigma_{f}=1$,
the segmentation results
are given in the fifth ($K_0 = 5$) and sixth ($K=100$) rows of Figure~\ref{fig4Images}. We further modified $\sigma_{f}$ to $\sqrt{10}$ in
our experiments, and found that the segmentation results are quite similar to
the setting $\sigma_{f}=1$ and thus not provided here. As is evident, using more clusters creates a finer segmentation, though the results are still similar. Subjectively, we see that LDDP outperforms $K$-means, while the two Pitman-Yor models do not have as clear a segmentation.


With most DP-based models, including the LDDP, the number of used clusters is expected to grow logarithmically with the number of observations~\cite{Teh10DP}. Therefore, it is not surprising that more clusters are used by the LDDP model when $K_0 =100$. We show this in Figure~\ref{figClustersize} for the four images considered. These plots give an ordered histogram of the number of pixels assigned to a cluster. We see that far fewer than 100 clusters contain data, highlighting the nonparametric aspect of the LDDP, but still more than the (possibly) desired number of segments. Therefore it is arguable that for image segmentation a nonparametric model is not ideal and $K_0$ should be set to a small number such as 5. We note that the LDDP can easily make this shift to parametric modeling as presented and derived above.

\subsection{Results with Ground Truth Segmentation}

We further compare our LDDP mixture model with the normalized cut
spectral clustering method~\cite{Shi00NC}, Pitman-Yor based models, and the GMM
using human-segmented images. We use the Rand
index~\cite{Unni07toe} to quantitatively evaluate the results. For
these images~\cite{Oliva01MS}, we know the number of true clusters
from the human segmentations and set $K_0$ to this number. While this is not possible in practice, we do this here for all algorithms as a head-to-head comparison. The
other settings are the same as the previous experiments.

\begin{wraptable}{r}{0.5\textwidth}\vspace{-20pt}
\footnotesize
\begin{tabular}{l|cccc}
 & rock & mountain & hut & building\\
\hline
Normalized cut  &78.63  &79.62 &78.35 &{\bf 81.86}\\
HPY  &58.61	&56.82	&56.02	&58.68\\
DPYP &59.01	&58.71	&58.57	&54.28\\
GMM  &{\bf 95.79}  & 80.98 & 81.63 &76.84\\
LDDP  &95.44  &{\bf 88.82} &{\bf 85.74} &74.00\\
 \hline
\end{tabular}
\caption{Comparison of Rand index values (\%).}\vspace{-10pt}
\label{tableRI}
\end{wraptable}

\begin{figure}[th!]
  \centering
  \includegraphics[width=.136\textwidth]{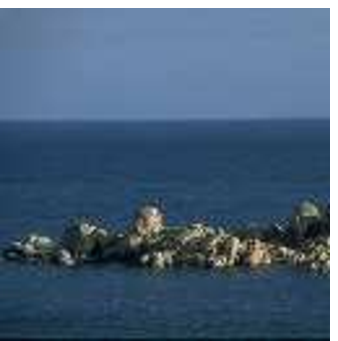}
  \includegraphics[width=.136\textwidth]{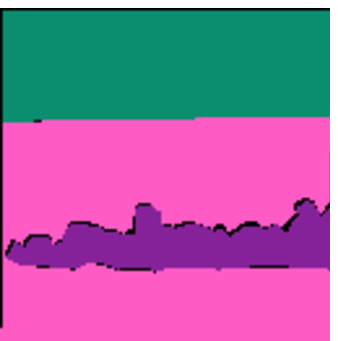}
    \includegraphics[width=.136\textwidth]{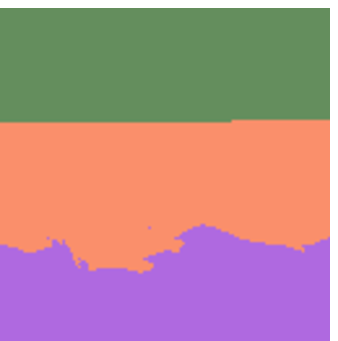}
\includegraphics[width=.136\textwidth]{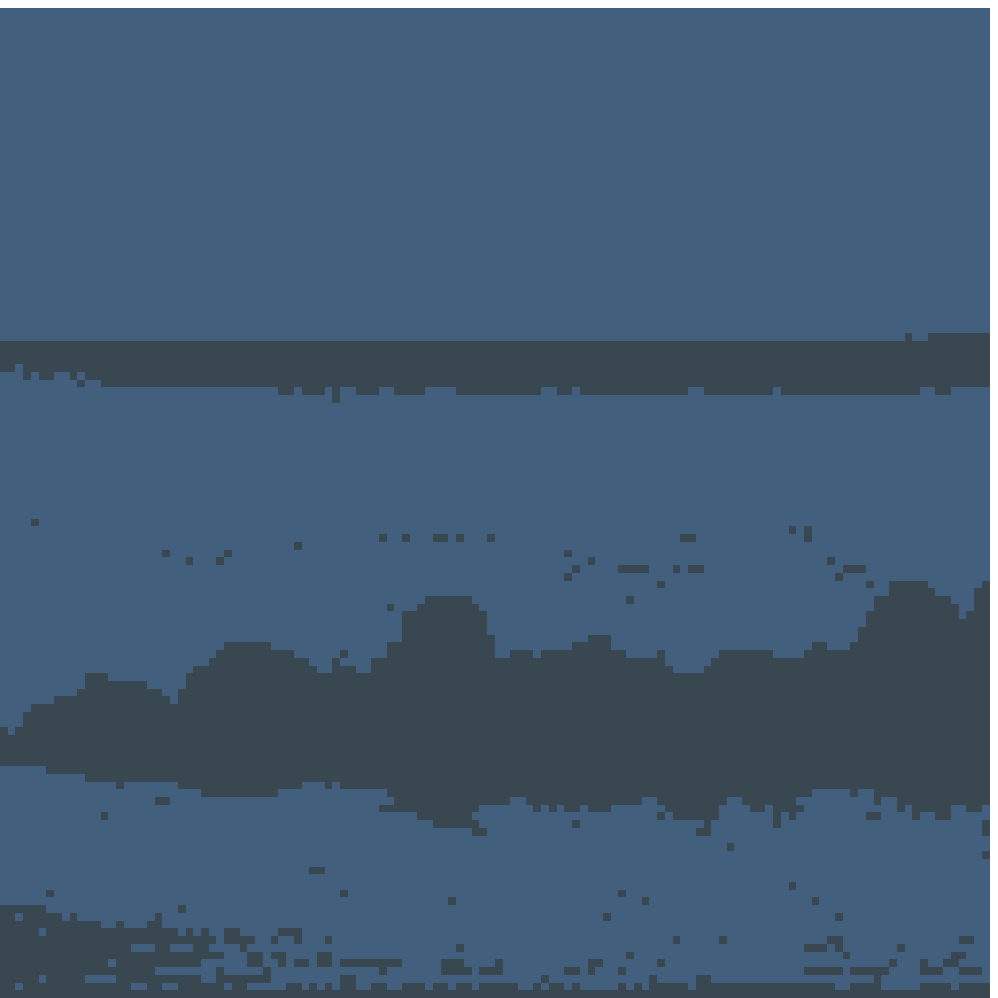}
 \includegraphics[width=.136\textwidth]{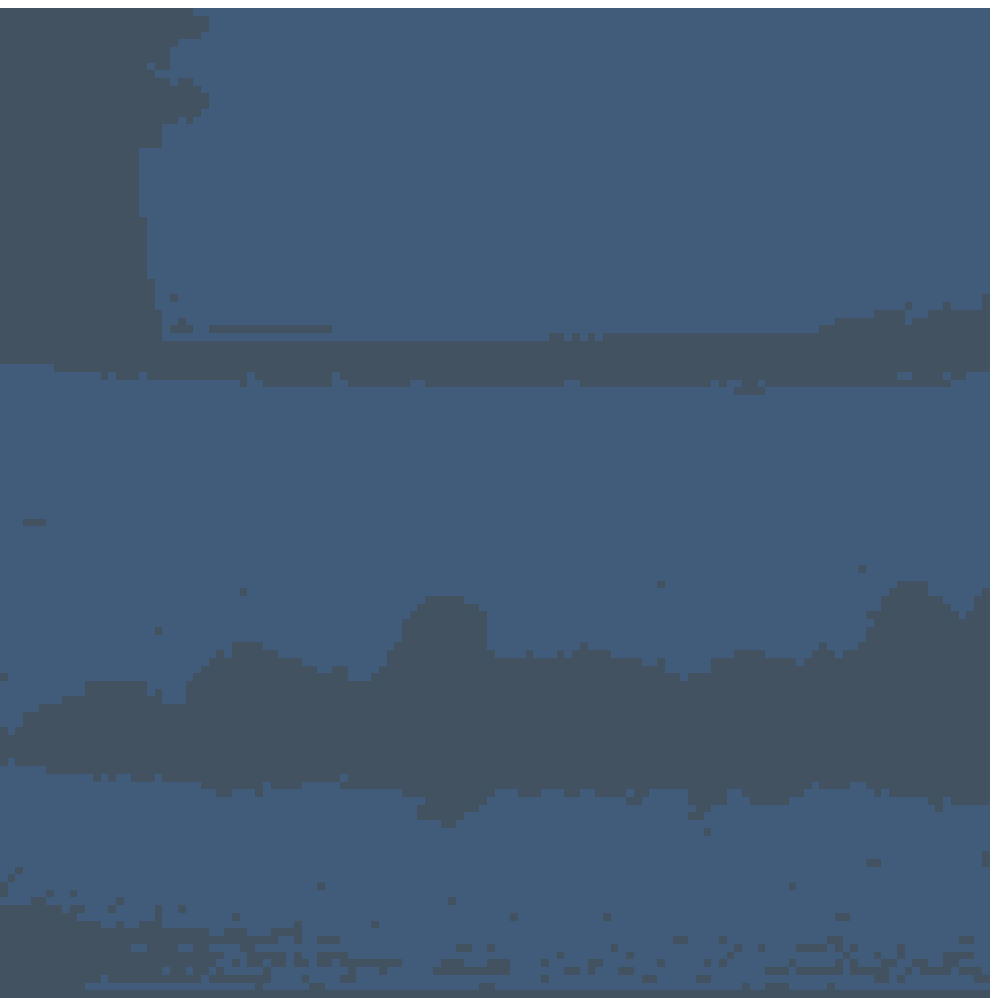}
  \includegraphics[width=.136\textwidth]{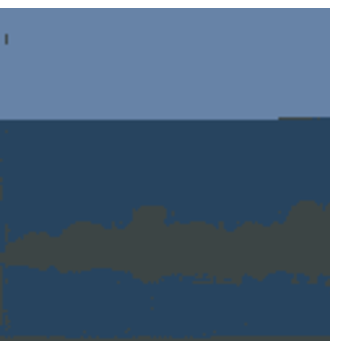}
  \includegraphics[width=.136\textwidth]{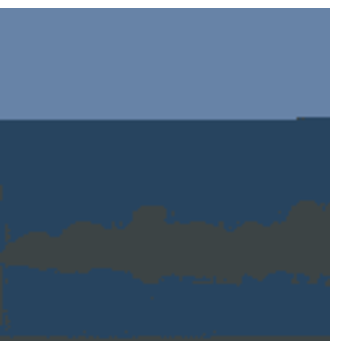}

  \includegraphics[width=.136\textwidth]{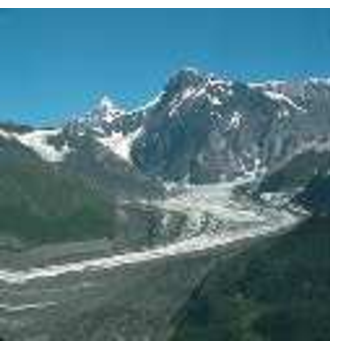}
    \includegraphics[width=.136\textwidth]{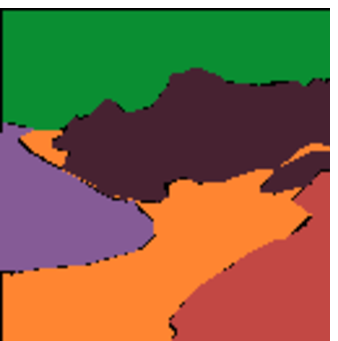}
  \includegraphics[width=.136\textwidth]{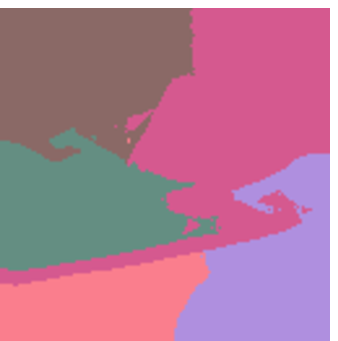}
  \includegraphics[width=.136\textwidth]{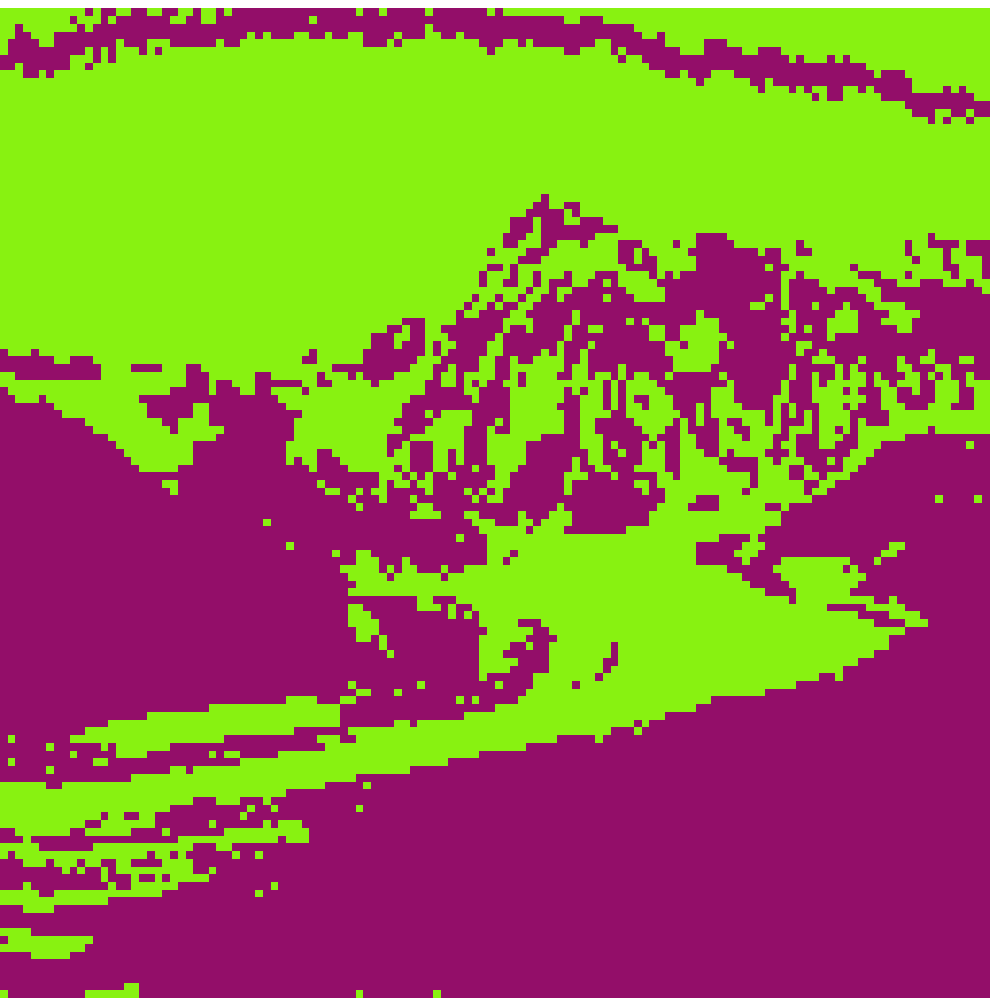}
 \includegraphics[width=.136\textwidth]{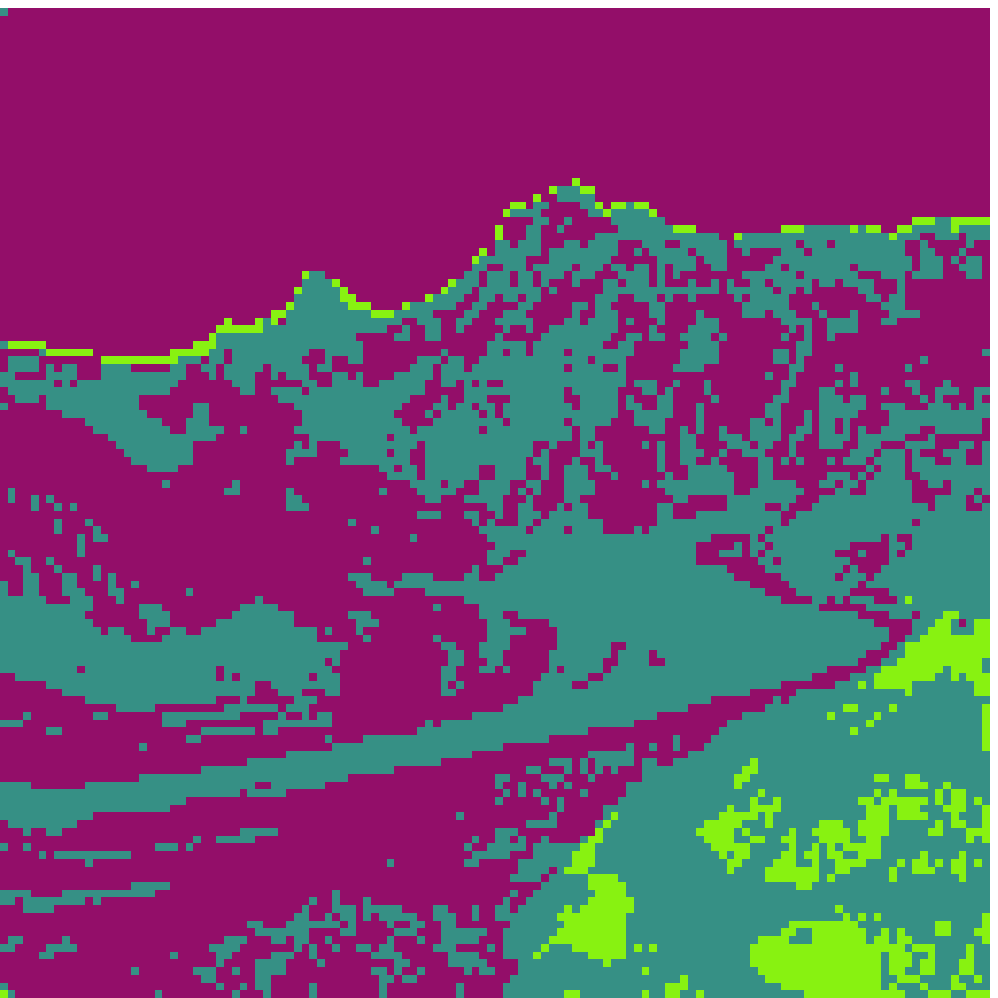}
  \includegraphics[width=.136\textwidth]{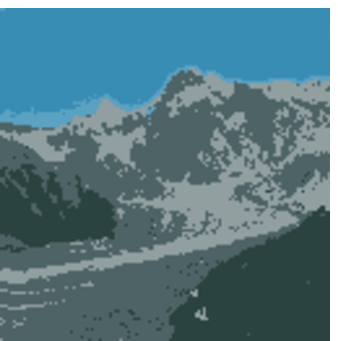}
  \includegraphics[width=.136\textwidth]{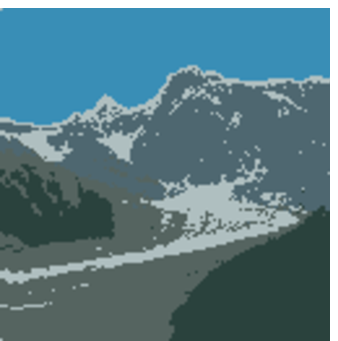}

  \includegraphics[width=.136\textwidth]{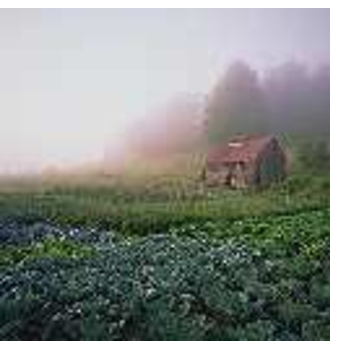}
    \includegraphics[width=.136\textwidth]{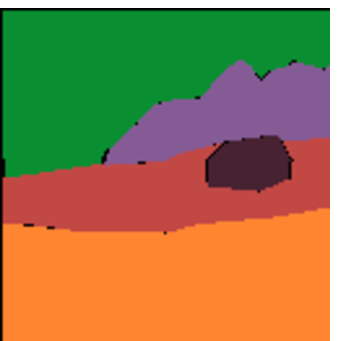}
  \includegraphics[width=.136\textwidth]{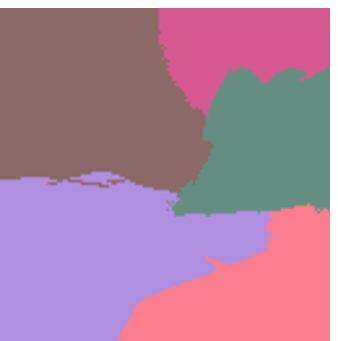}
 \includegraphics[width=.136\textwidth]{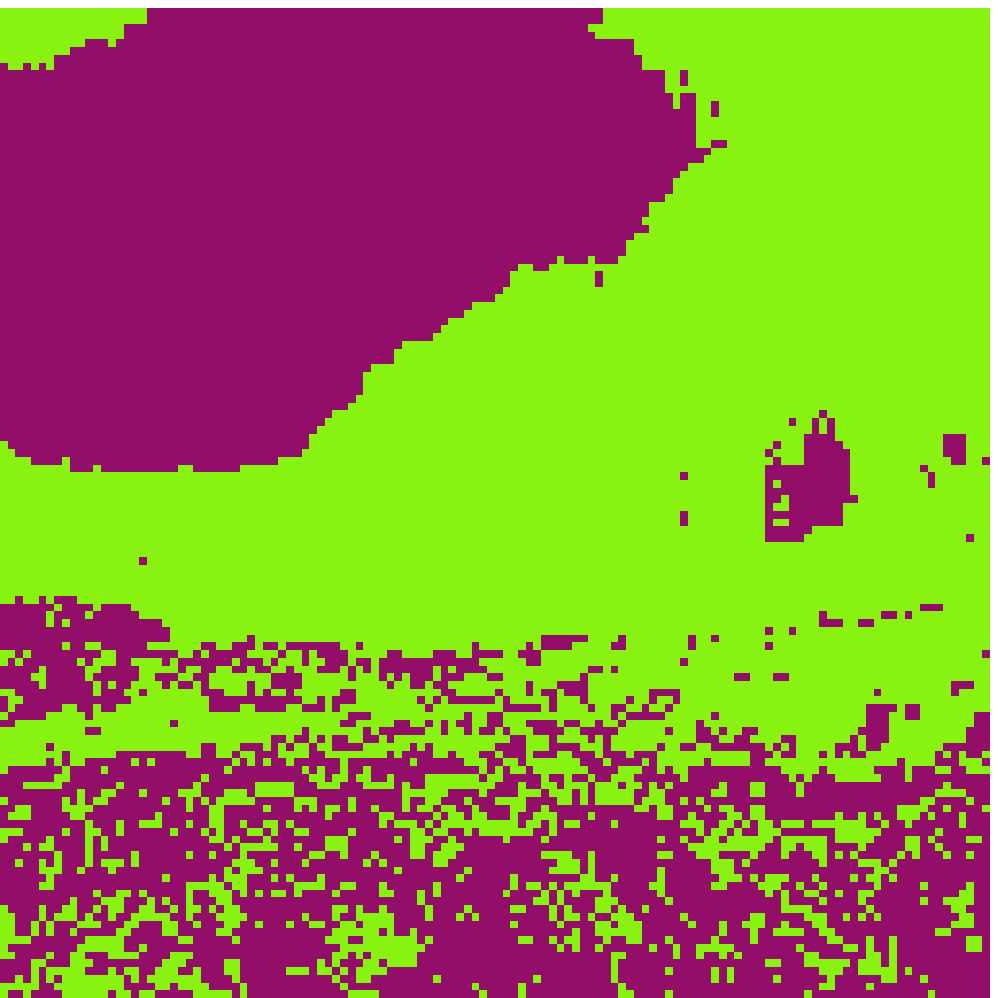}
 \includegraphics[width=.136\textwidth]{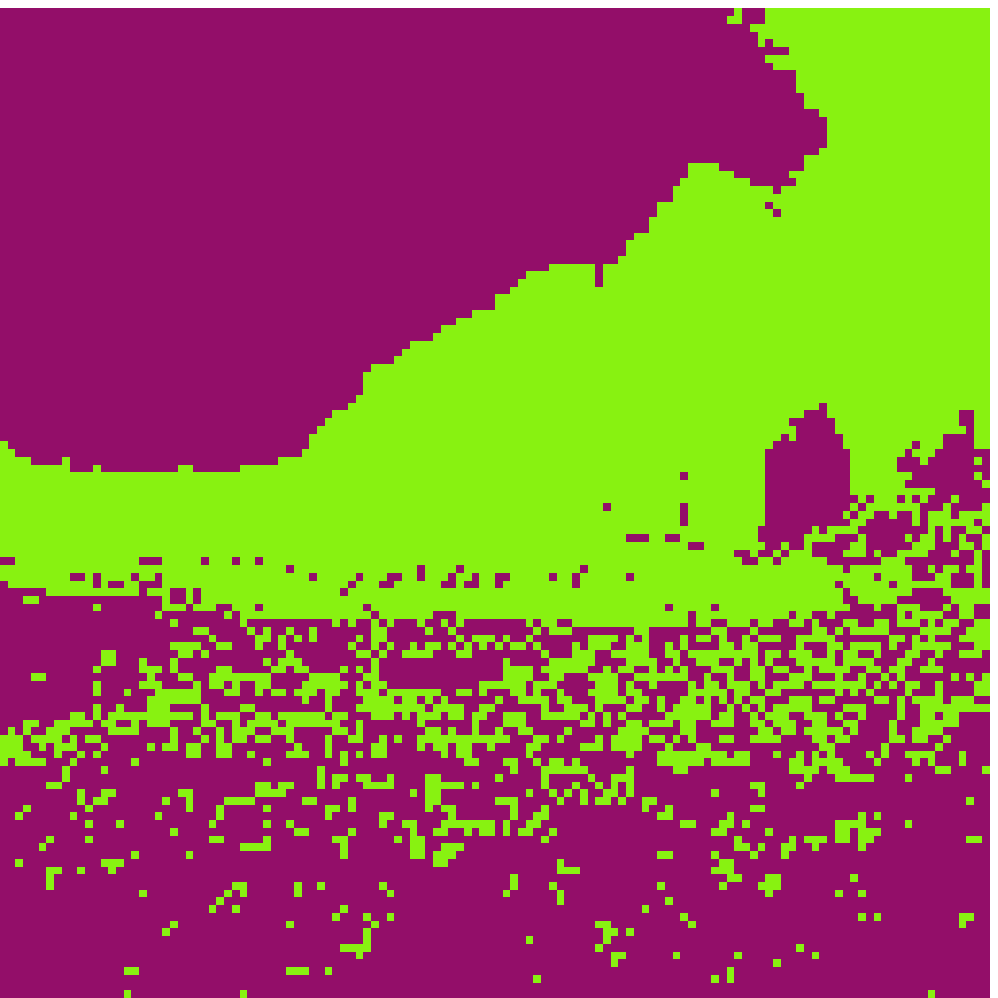}
  \includegraphics[width=.136\textwidth]{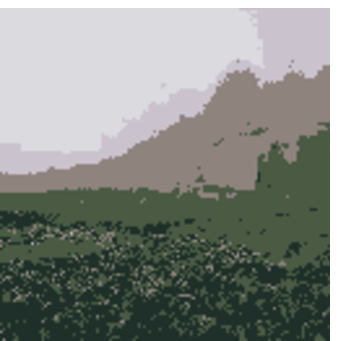}
  \includegraphics[width=.136\textwidth]{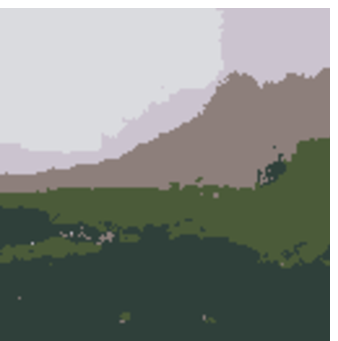}\vspace{-5pt}

  \subfigure[Orig.]{\includegraphics[width=.136\textwidth]{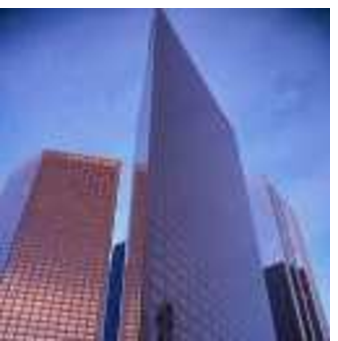}}
  \subfigure[Human]{\includegraphics[width=.136\textwidth]{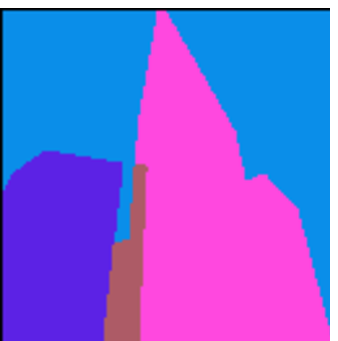}}
  \subfigure[NrmCt]{\includegraphics[width=.136\textwidth]{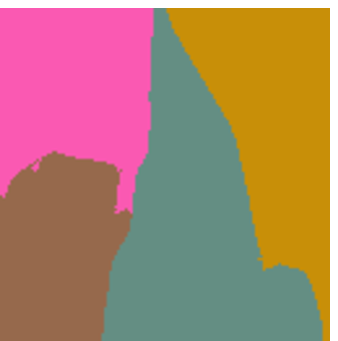}}
 \subfigure[HPY]{\includegraphics[width=.136\textwidth]{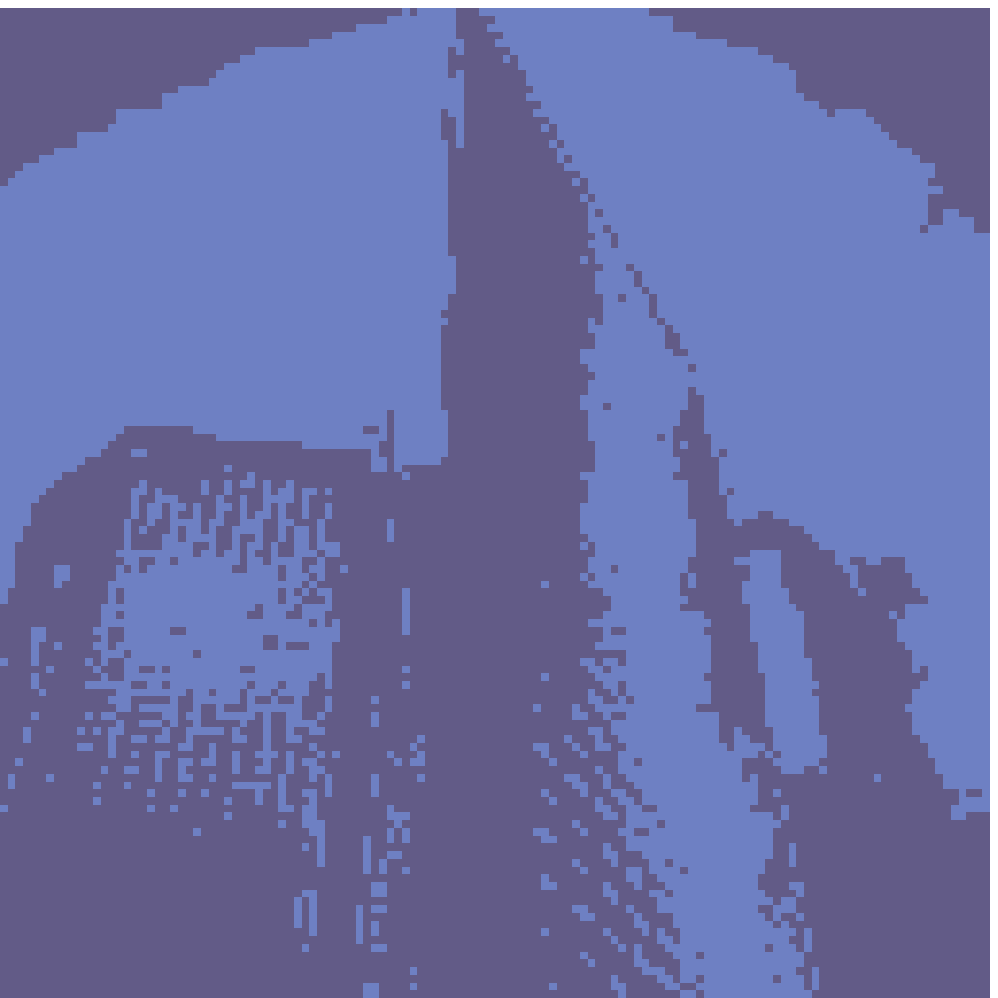}}
 \subfigure[DPYP]{\includegraphics[width=.136\textwidth]{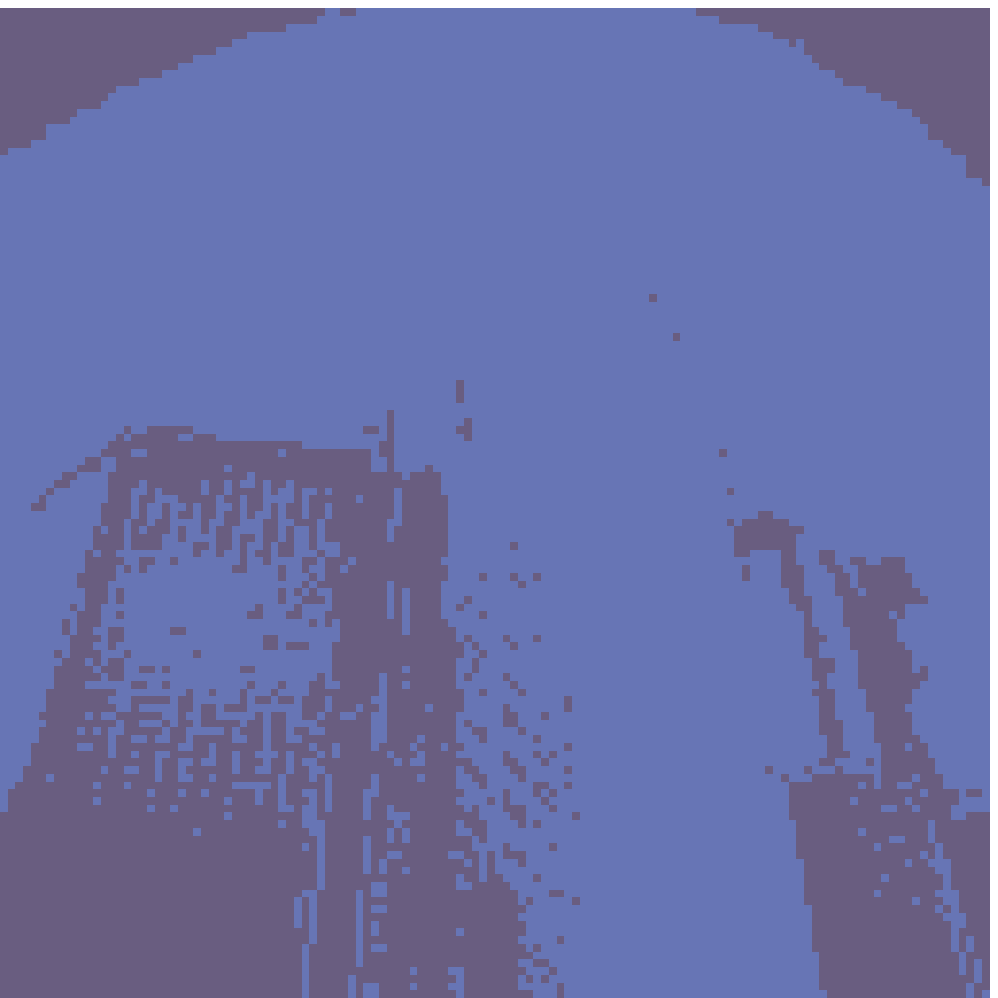}}
  \subfigure[GMM]{\includegraphics[width=.136\textwidth]{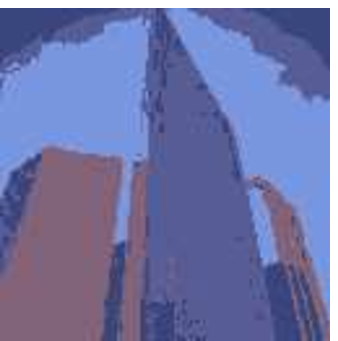}}
  \subfigure[LDDP]{\includegraphics[width=.136\textwidth]{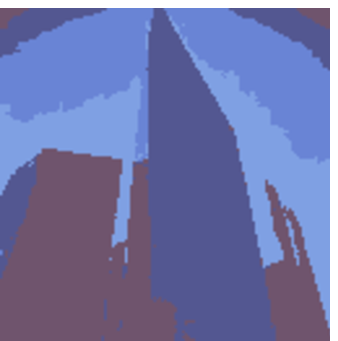}}
  \caption{Example segmentations when ground truth is known. The number of clusters is set to ground truth for all experiments to facilitate a head-to-head comparison of the modeling structure. LDDP performs better than other Bayesian methods and comparable to normalized cuts. We see the improvement of LDDP over GMM as a result of the added Gaussian processes.}
  \label{fig4ImageswGT}
\end{figure}

The original images, ground truth and segmentation results are all shown
in Figure~\ref{fig4ImageswGT}.
Although the normalized cut spectral clustering method also leverages
the spatial location of the pixels,
it  is implicitly biased towards regions of equal size,
 as we can see in the third column of Figure~\ref{fig4ImageswGT}. The LDDP appears to perform qualitatively better than other Bayesian methods. We particularly note the improvement over the GMM, a special case of LDDP, which is due to the addition of Gaussian processes to each cluster.

The Rand index is a standard quantitative measurement of the
similarity between a segmentation and the ground truth. The
corresponding Rand index values for the images segmentation results
are given in Table~\ref{tableRI}. These results indicate the overall competitiveness of LDDP for segmentation. We again highlight the general improvement over other general Bayesian methods, which like LDDP are also applicable to a broader set of modeling problems.

\section{Conclusion}
We have proposed location dependent Dirichlet
processes (LDDP), a general mixture modeling framework for clustering data using additional location information. We derived a general variational inference algorithm for both parametric and approximately nonparametric settings. We presented a case study of a Gaussian LDDP for an image segmentation task where we saw competitive results. Future research will focus on exploring more applications of the proposed framework beyond the current image data.

\bibliographystyle{splncs}
\bibliography{sample}
\end{document}